%% file: main.tex
\documentclass[runningheads]{template/llncs}
\usepackage{graphicx, amsmath, amssymb, color}
\usepackage{subfig, microtype, mwe, booktabs, pifont, arydshln, url, appendix}
\usepackage[dvipsnames]{xcolor}
\usepackage{definitions}
\usepackage[dvipsnames]{xcolor}

\arxivcopy 

\begin{reviewonly}
    \newcommand{\Appendix}{supplementary material}
\end{reviewonly}
\begin{finalonly}
    \newcommand{\Appendix}{Appendix}
\end{finalonly}

\begin{reviewonly}
    \usepackage{template/ruler}
    \usepackage[width=122mm,left=12mm,paperwidth=146mm,height=193mm,top=12mm,paperheight=217mm]{geometry}
\end{reviewonly}
\begin{arxivonly}
    \usepackage[width=122mm,left=12mm,paperwidth=146mm,height=193mm,top=12mm,paperheight=217mm]{geometry}
    \usepackage[pagebackref=true,breaklinks=true,colorlinks,bookmarks=false,linkcolor=NavyBlue]{hyperref}
\end{arxivonly}

\def\ECCVSubNumber{849}

\begin{arxivonly}
    \usepackage{times}
\end{arxivonly}

\begin{document}

\pagestyle{headings}
\mainmatter


\title{TIDE: A General Toolbox for Identifying Object Detection Errors}

\begin{reviewonly}
	\titlerunning{ECCV-20 submission ID \ECCVSubNumber} 
	\authorrunning{ECCV-20 submission ID \ECCVSubNumber} 
	\author{Anonymous ECCV submission}
	\institute{Paper ID \ECCVSubNumber}
\end{reviewonly}

\begin{finalonly}
	\titlerunning{TIDE: A General Toolbox for Identifying Object Detection Errors}
	\author{Daniel Bolya \and 
			Sean Foley \and 
			James Hays \and 
			Judy Hoffman} 
	\authorrunning{D. Bolya et al.}
	\institute{Georgia Institute of Technology}
\end{finalonly}

\maketitle

\input{sections/0_abstract}

\input{sections/1_intro}

\input{sections/2_toolkit}

\input{sections/3_analysis}

\input{sections/4_conclusion}


\bibliographystyle{template/splncs04}
\bibliography{main}

\begin{arxivonly}
\input{sections/5_appendix}
\end{arxivonly}

\end{document}

%% file: sections/0_abstract.tex
\begin{abstract}
	We introduce TIDE, a framework and associated toolbox\footnote{\small{\url{https://dbolya.github.io/tide/}}} for analyzing the sources of error in object detection and instance segmentation algorithms. Importantly, our framework is applicable across datasets and can be applied directly to output prediction files without required knowledge of the underlying prediction system. Thus, our framework can be used as a drop-in replacement for the standard mAP computation while providing a comprehensive analysis of each model's strengths and weaknesses. We segment errors into six types and, crucially, are the first to introduce a technique for measuring the contribution of each error in a way that isolates its effect on overall performance. We show that such a representation is critical for drawing accurate, comprehensive conclusions through in-depth analysis across 4 datasets and 7 recognition models.
	
\keywords{Error Diagnosis, Object Detection, Instance Segmentation}
\end{abstract}

%% file: sections/1_intro.tex
\section{Introduction}
Object detection and instance segmentation are fundamental tasks in computer vision, with applications ranging from self-driving cars \cite{cordts2016cityscapes} to tumor detection \cite{dong2017tumor}. Recently, the field of object detection has rapidly progressed, thanks in part to competition on challenging benchmarks, such as CalTech Pedestrians \cite{dollar2009caltechpeds}, 
Pascal \cite{everingham2010pascal}, COCO \cite{lin2014coco}, Cityscapes \cite{cordts2016cityscapes}, and LVIS \cite{gupta2019lvis}. Typically, performance on these benchmarks is summarized by one number: mean Average Precision ($mAP$). 

However, $mAP$ suffers from several shortcomings, not the least of which is its complexity. It is defined as the area under the precision-recall curve for detections at a specific intersection-over-union (IoU) threshold with a correctly classified ground truth (GT), averaged over all classes. Starting with  COCO~\cite{lin2014coco}, it became standard to average $mAP$ over 10 IoU thresholds (interval of 0.05) to get a final $mAP^{0.5:0.95}$. The complexity of this metric poses a particular challenge when we wish to analyze errors in our detectors, as error types become intertwined, making it difficult to gauge how much each error type affects $mAP$.

Moreover, by optimizing for $mAP$ alone, we may be inadvertently leaving out the relative importance of error types that can vary between applications. For instance, in tumor detection, correct classification arguably matters more than box localization; the existence of the tumor is essential, but the precise location may be manually corrected. In contrast, precise localization may be critical for robotic grasping where even slight mislocalizations can lead to faulty manipulation. Understanding how these sources of error relate to overall $mAP$ is crucial to designing new models and choosing the proper model for a given task.

Thus we introduce TIDE, a general Toolkit for Identifying Detection and segmentation Errors, in order to address these concerns. We argue that a complete toolkit should: 1.) compactly summarize error types, so comparisons can be made at a glance; 2.) fully isolate the contribution of each error type, such that there are no confounding variables that can affect conclusions; 3.) not require dataset-specific annotations, to allow for comparisons across datasets; 4.) incorporate all the predictions of a model, since considering only a subset hides information; 5.) allow for finer analysis as desired, so that the sources of errors can be isolated.

\input{figures/7_toolkit_intro_comparison}

\paragraph{Why we need a new analysis toolkit.} 
Many works exist to analyze the errors in object detection and instance segmentation \cite{hosang2014proposals,zhu2015diagnosing,divvala2009context,kabra2015neighbors,pepik2015holding}, but only a few provide a useful summary of all the errors in a model \cite{hoiem2012diagnosing,COCOtoolkit,borji2019empirical}, and none have all the desirable properties listed above.

Hoiem et al. introduced the foundational work for summarizing errors in object detection \cite{hoiem2012diagnosing}, however their summary only explains false positives (with false negatives requiring separate analysis), and it depends on a hyperparameter $N$ to control how many errors to consider, thus not fulfilling (4). Moreover, to use this summary effectively, this $N$ needs to be swept over which creates 2d plots that are difficult to interpret (see error analysis in \cite{girshick2014rcnn,liu2016ssd}), and thus in practice only partially addresses (1). Their approach also doesn't fulfill (3) because their error types require manually defined superclasses which are not only subjective, but difficult to meaningfully define for datasets like LVIS \cite{gupta2019lvis} with over 1200 classes. Finally, it only partially fulfills (2) since the classification errors are defined such that if the detection is both mislocalized and misclassified it will be considered as misclassified, limiting the effectiveness of conclusions drawn from classification and localization error.

The COCO evaluation toolkit \cite{COCOtoolkit} attempts to update Hoiem et al.'s work by representing errors in terms of their effect on the precision-recall curve (thus tying them closer to $mAP$). This allows them to use all detections at once (4), since the precision recall curve implicitly weights each error based on its confidence. However, the COCO toolkit generates 372 2d plots, each with 7 precision-recall curves, which requires a significant amount of time to digest and thus makes it difficult to compactly compare models (1). Yet, perhaps the most critical issue is that the COCO eval toolkit computes errors progressively which we show drastically misrepresents the contribution of each error (2), potentially leading to incorrect conclusions (see Sec.~\ref{sec:progressive_error}). Finally, the toolkit requires manual annotations that exist for COCO but not necessarily for other datasets (3).

As concurrent work, \cite{borji2019empirical} attempts to find an upper bound for $AP$ on these datasets and in the process addresses certain issues with the COCO toolkit. However, this work still bases their error reporting on the same progressive scheme that the COCO toolkit uses, which leads them to the dubious conclusion that background error is significantly more important all other types (see Fig.~\ref{fig:4_progressive_errors}). As will be described in detail later, to draw reliable conclusions, it is essential that our toolkit work towards isolating the contribution of each error type (2). 

\paragraph{Contributions}
In our work, we address all 5 goals and provide a compact, yet detailed {\it summary} of the errors in object detection and instance segmentation. Each error type can be represented as a single meaningful number (1), making it compact enough to fit in ablation tables (see Tab.~\ref{tab:6_example_ablation}), incorporates all detections (4), and doesn't require any extra annotations (3). We also weight our errors based on their effect on overall performance while carefully avoiding the confounding factors present in $mAP$ (2). And while we prioritize ease of interpretation, our approach is modular enough that the same set of errors can be used for more fine-grained analysis (5). The end result is a compact, meaningful, and expressive set of errors that is applicable across models, datasets, and even tasks.

We demonstrate the value of our approach by comparing several recent CNN-based object detectors and instance segmenters across several datasets. We explain how to incorporate the summary into ablation studies to quantitatively justify design choices. We also provide an example of how to use the summary of errors to guide more fine-grained analysis in order to identify specific strengths or weaknesses of a model.

We hope that this toolkit can form the basis of analysis for future work, lead model designers to better understand weaknesses in their current approach, and allow future authors to quantitatively and compactly justify their design choices. To this end, full toolkit code is released at {\small \url{https://dbolya.github.io/tide/}} and opened to the community for future development.

%% file: figures/7_toolkit_intro_comparison.tex
\begin{table}[t]
    \centering
    
    \newcommand{\yes}{\ding{52}}
    \newcommand{\no}{\ding{55}}
    \newcommand{\maybe}{\ding{81}}
    
    \caption{\textbf{Comparison to Other Toolkits.} We compare our desired features between existing toolkits and ours. \yes{} indicates a toolkit has the feature, \maybe{} indicates that it partially does, and \no{} indicates that it doesn't. }
    
    \begin{smalltable}{l c c c c} \toprule
        Feature & Hoiem \cite{hoiem2012diagnosing} & COCO \cite{COCOtoolkit} & UAP \cite{borji2019empirical} & TIDE (Ours) \\
        \midrule
        Compact Summary of Error Types & \maybe &  \no & \yes & \yes \\
        Isolates Error Contribution    & \maybe &  \no &  \no & \yes \\
        Dataset Agnostic               &    \no &  \no & \yes & \yes \\
        Uses All Detections            &    \no & \yes & \yes & \yes \\
        Allows for deeper analysis     &   \yes & \yes & \yes & \yes \\
        \bottomrule
    \end{smalltable}
    
    \label{fig:7_toolkit_intro_comparison}
\end{table}

%% file: sections/2_toolkit.tex
\section{The Tools}

Object detection and instance segmentation primarily use one metric to judge performance: mean Average Precision ($mAP$). While $mAP$ succinctly summarizes the performance of a model in one number, disentangling errors in object detection and instance segmentation from $mAP$ is difficult: a false positive can be a duplicate detection, misclassification, mislocalization, confusion with background, or even both a misclassification and mislocalization. Likewise, a false negative could be a completely missed ground truth, or the potentially correct prediction could have just been misclassified or mislocalized. These error types can have hugely varying effects on $mAP$, making it tricky to diagnose problems with a model off of $mAP$ alone.

We could categorize all these types of errors, but it's not entirely clear how to weight their relative importance. Hoiem et al.~\cite{hoiem2012diagnosing} weight false positives by their prevalence in the top $N$ most confident errors and consider false negatives separately. However, this ignores the effect many low scoring detections could have (so effective use of it requires a sweep over $N$), and it doesn't allow comparison between false positives and false negatives.

There is one easy way to determine the importance of a given error to overall $mAP$, however: simply fix that error and observe the resulting change in $mAP$. Hoiem et al.\ briefly explored this method for certain false positives but didn't base their analysis off of it. This is also similar to how the COCO eval toolkit \cite{COCOtoolkit} plots errors, with one key difference: the COCO implementation computes the errors {\it progressively}. That is, it observes the change in $mAP$ after fixing one error, but keep those errors fixed for the next error. This is nice because at the end result is trivially 100 $mAP$, but we find that fixing errors progressively in this manner is misleading and may lead to false conclusions (see Sec.~\ref{sec:progressive_error}).

So instead, we define errors in such a way that fixing all errors will still result in 100 $mAP$, but we weight each error individually starting from the original model's performance. This retains the nice property of including confidence and false negatives in the calculation, while keeping the magnitudes of each error type comparable.

\subsection{Computing mAP}
\label{sec:computing_map}
    
    Before defining error types, we focus our attention on the definition of $mAP$ to understand what may cause it to degrade. To compute $mAP$, we are first given a list of predictions for each image by the detector. Each ground truth in the image is then matched to at most one detection. To qualify as a positive match, the detection must have the same class as the ground truth and an IoU overlap greater than some threshold, $t_f$, which we will consider as 0.5 unless otherwise specified. If multiple detections are eligible, the one with the highest overlap is chosen to be true positive while all remaining are considered false positives.
    
    Once each detection has matched with a ground truth (true positive) or not (false positive), all detections are collected from every image in the dataset and are sorted by descending confidence. Then the cumulative precision and recall over all detections is computed as:
    \begin{equation}
        P_c = \frac{TP_c}{TP_c + FP_c} \qquad R_c = \frac{TP_c}{N_{GT}}
    \label{eq:precision_and_recall}
    \end{equation}
    where for all detections with confidence $\geq c$, $P_c$ denotes the precision, $R_c$ recall, $TP_c$ the number of true positives, and $FP_c$ the number of false positives. $N_{GT}$ denotes the number of GT examples in the current class.
    
    Then, precision is interpolated such that $P_c$ decreases monotonically, and $AP$ is computed as a integral under the precision recall curve (approximated by a fixed-length Riemann sum). Finally, $mAP$ is defined as the average $AP$ over all classes. In the case of COCO \cite{lin2014coco}, $mAP$ is averaged over all IoU thresholds between $0.50$ and $0.95$ with a step size of $0.05$ to obtain $mAP^{0.5:0.95}$.

\input{figures/2_error_types_definition}

\subsection{Defining Error Types}
    
    Examining this computation, there are 3 places our detector can affect $mAP$: outputting false positives during the matching step, not outputting true positives (i.e., false negatives) for computing recall, and having incorrect calibration (i.e., outputting a higher confidence for a false positive then a true positive).
    
    \subsubsection{Main Error Types}
    
    In order to create a meaningful distribution of errors that captures the components of $mAP$, we bin all false positives and false negatives in the model into one of 6 types (see Fig.~\ref{fig:2_error_type_definition}). Note that for some error types (classification and localization), a false positive can be paired with a false negative. We will use $IoU_\text{max}$ to denote a false positive's maximum IoU overlap with a ground truth of the given category. The foreground IoU threshold is denoted as $t_f$ and the background threshold is denoted as $t_b$, which are set to 0.5 and 0.1 (as in \cite{hoiem2012diagnosing}) unless otherwise noted.
    \begin{enumerate}
        \item \textbf{Classification Error}:  $IoU_\text{max} \geq t_\text{f}$ for GT of the {\it incorrect} class (i.e., localized correctly but classified incorrectly).
        \item \textbf{Localization Error}: $t_\text{b} \leq IoU_\text{max} \leq t_\text{f}$ for GT of the {\it correct} class (i.e., classified correctly but localized incorrectly).
        \item \textbf{Both Cls and Loc Error}: $t_\text{b} \leq IoU_\text{max} \leq t_\text{f}$ for GT of the {\it incorrect} class (i.e., classified incorrectly \textit{and} localized incorrectly).
        \item \textbf{Duplicate Detection Error}: $IoU_\text{max} \geq t_\text{f}$ for GT of the {\it correct} class but another higher-scoring detection already matched that GT (i.e., would be correct if not for a higher scoring detection).
        \item \textbf{Background Error}: $IoU_\text{max} \leq t_\text{b}$ for all GT (i.e., detected background as foreground).
        \item \textbf{Missed GT Error}: All undetected ground truth (false negatives) not already covered by classification or localization error.
    \end{enumerate}
    
    This differs from \cite{hoiem2012diagnosing} in a few important ways. First, we combine both \texttt{sim} and \texttt{other} errors into one classification error, since Hoiem et al.'s \texttt{sim} and \texttt{other} require manual annotations that not all datasets have and analysis of the distinction can be done separately. Then, both classification errors in \cite{hoiem2012diagnosing} are defined for all detections with $IoU_\text{max} \geq t_b$, even if $IoU_\text{max} < t_f$. This confounds localization and classification errors, since using that definition, detections that are both mislocalized and misclassified are considered class errors. Thus, we separate these detections into their own category.

    \subsubsection{Weighting the Errors}
    
    Just counting the number of errors in each bin is not enough to be able to make direct comparisons between error types, since a false positive with a lower score has less effect on overall performance than one with a higher score. Hoiem et al. \cite{hoiem2012diagnosing} attempt to address this by considering the top $N$ highest scoring errors, but in practice $N$ needed to be swept over to get the full picture, creating 2d plots that are hard to interpret (see the analysis in \cite{girshick2014rcnn,liu2016ssd}).
    
    Ideally, we'd like one comprehensive number that represents how each error type affects overall performance of the model. In other words, for each error type we'd like to ask the question, how much is this category of errors holding back the performance of my model? In order to answer that question, we can consider what performance of the model would be if it didn't make that error and use how that changed $mAP$.
    
    To do this, for each error we need to define a corresponding ``oracle'' that fixes that error. For instance, if an oracle $o \in \mathcal{O}$ described how to change some false positives into true positives, we could call the $AP$ computed after applying the oracle as $AP_o$ and then compare that to the vanilla $AP$ to obtain that oracle's (and corresponding error's) effect on performance:
    \begin{equation}
        \Delta AP_o = AP_o - AP
    \label{eq:delAP_defn}
    \end{equation}
    We know that we've covered all errors in the model if applying all the oracles together results in 100 $mAP$. In other words, given oracles $\mathcal{O} = \{o_1, \ldots, o_n\}$:
    \begin{equation}
        AP_{o_1, \ldots, o_n} = 100 \qquad AP + \Delta AP_{o_1, \ldots, o_n} = 100
    \label{eq:delAP_total}
    \end{equation}
    Referring back to the definition of $AP$ in Sec.~\ref{sec:computing_map}, to satisfy Eq.~\ref{eq:delAP_total} the oracles used together must fix all false positives and false negatives.
    
    Considering this, we define the following oracles for each of the main error types described above:
    \begin{enumerate}
        \item \textbf{Classification Oracle}: Correct the class of the detection (thereby making it a true positive). If a duplicate detection would be made this way, suppress the lower scoring detection.
        \item \textbf{Localization Oracle}: Set the localization of the detection to the GT's localization (thereby making it a true positive). Again, if a duplicated detection would be made this way, suppress the lower scoring detection.
        \item \textbf{Both Cls and Loc Oracle}: Since we cannot be sure of which GT the detector was attempting to match to, just suppress the false positive detection.
        \item \textbf{Duplicate Detection Oracle}: Suppress the duplicate detection.
        \item \textbf{Background Oracle}: Suppress the hallucinated background detection.
        \item \textbf{Missed GT Oracle}: Reduce the number of GT ($N_{GT}$) in the $mAP$ calculation by the number of missed ground truth. This has the effect of stretching the precision-recall curve over a higher recall, essentially acting as if the detector was equally as precise on the missing GT. The alternative to this would be to add new detections, but it's not clear what the score should be for that new detection such that it doesn't introduce confounding variables. We discuss this choice further in the \Appendix.
    \end{enumerate}

    \subsubsection{Other Error Types}
    
    While the previously defined types fully account for all error in the model, how the errors are defined doesn't clearly delineate false positive and negative errors (since cls, loc, and missed errors can all capture false negatives). There are cases where a clear split would be useful, so for those cases we define two separate error types by the oracle that would address each:
    \begin{enumerate}
        \item \textbf{False Positive Oracle}: Suppress all false positive detections.
        \item \textbf{False Negative Oracle}: Set $N_{GT}$ to the number of true positive detections.
    \end{enumerate}
    Both of these oracles together account for 100 $mAP$ like the previous 6 oracles do, but they bin the errors in a different way.

    \subsection{Limitations of Computing Errors Progressively}
    \label{sec:progressive_error}
    
        \input{figures/4_progressive_error}

    Note that we are careful to compute errors \textit{individually} (i.e., each $\Delta AP$ starts from the vanilla $AP$ with no errors fixed). Other approaches \cite{COCOtoolkit,borji2019empirical} compute their errors \textit{progressively} (i.e., each $\Delta AP$ starts with the last error fixed, such that fixing the last error results in 100 $AP$). While we ensure that applying all oracles together also results in 100 $AP$, we find that a progressive $\Delta AP$ misrepresents the weight of each error type and is strongly biased toward error types \emph{fixed last}.
    
    To make this concrete, we can define progressive error $\Delta AP_{a \mid b}$ to be the change in $AP$ from applying oracle $a$ given that you've already applied oracle $b$:
    \begin{equation}
        \Delta AP_{a \mid b} = AP_{a,b} - AP_b
    \label{eq:progressive_AP_defn}
    \end{equation}
    Then, computing errors progressively amounts to setting the importance of error $i$ to $\Delta AP_{o_i \mid o_1, \ldots, o_{i-1}}$. This is problematic for two reasons: the definition of precision includes false positives in the {\it denominator}, meaning that if you start with fewer false positives (as would be the case when having fixed most false positives already), the change in precision will be {\it much higher}. Furthermore, any changes in recall (e.g., by fixing localization or classification errors) amplifies the effect of precision on $mAP$, since the integral now has more area.

    We show this empirically in Fig.~\ref{fig:4_progressive_errors}, where Fig.~\ref{subfig:4_progressive_original} displays the original COCO eval style PR curves, while Fig.~\ref{subfig:4_progressive_swapped} simply swaps the order that background and classification error are computed. Just computing background first leads to an incredible decrease in the prevalence of its contribution (given by the area of the shaded region), meaning that the true weight of background error is likely much less than COCO eval reports. This makes it difficult to draw factual conclusions from analysis done this way.
    
    Moreover, computing errors progressively doesn't make intuitive sense. When using these errors, you'd be attempting to address them individually, one at a time. There will never be an opportunity to correct {\it all} localization errors, and then start addressing the classification errors---there will always be some amount of error in each category left over after improving the method, so observing $AP_{a \mid b}$ isn't useful, because there is no state where you're starting with $AP_b$.
    
    For these reasons, we entirely avoid computing errors progressively.

%% file: figures/2_error_types_definition.tex
\begin{figure}[t]
    \begin{center}
    
    \begin{tabular}{c: c: c: c: c: c}
         \small{\textbf{Cls}} & \small{\textbf{Loc}} &
        \small{\textbf{Cls+Loc}} & 
        \small{\textbf{Duplicate}} & 
        \small{\textbf{Bkgd}} & 
        \small{\textbf{Missed}}\\
        \midrule
         \includegraphics[width=.16\linewidth]{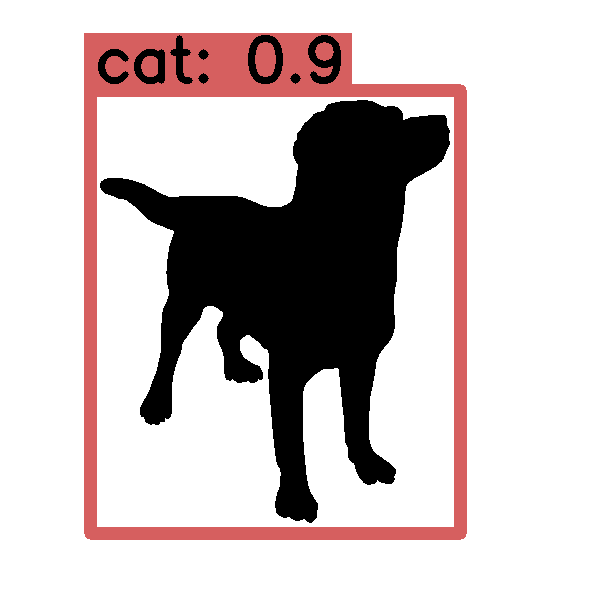} &  
        \includegraphics[width=.16\linewidth]{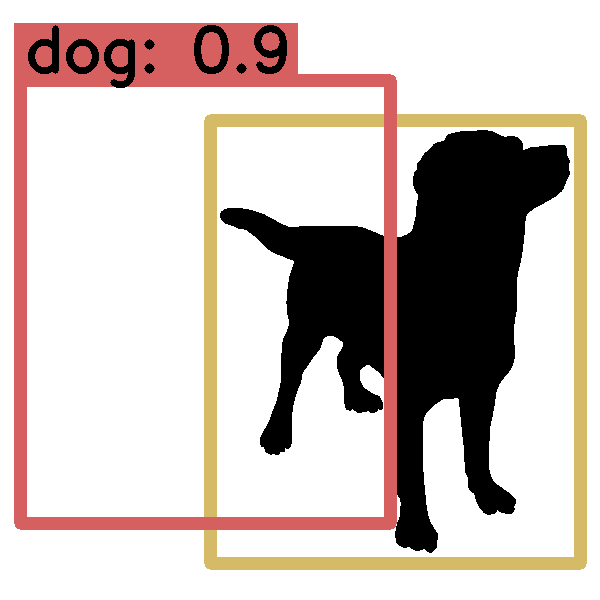} &
        \includegraphics[width=.16\linewidth]{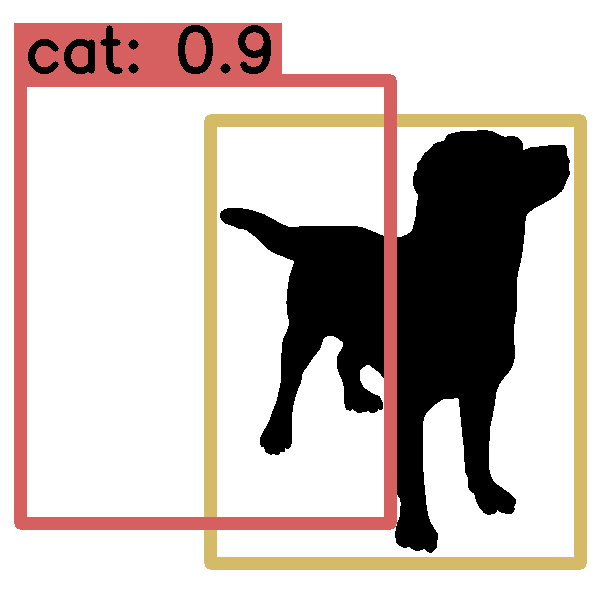} &
        \includegraphics[width=.16\linewidth]{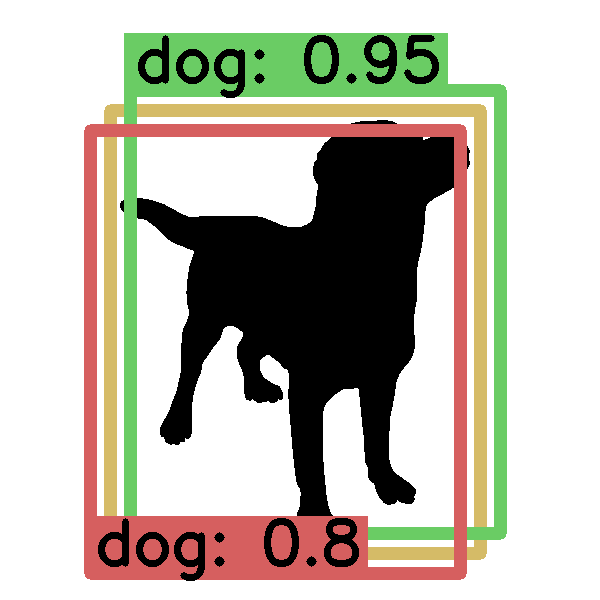} &
        \includegraphics[width=.16\linewidth]{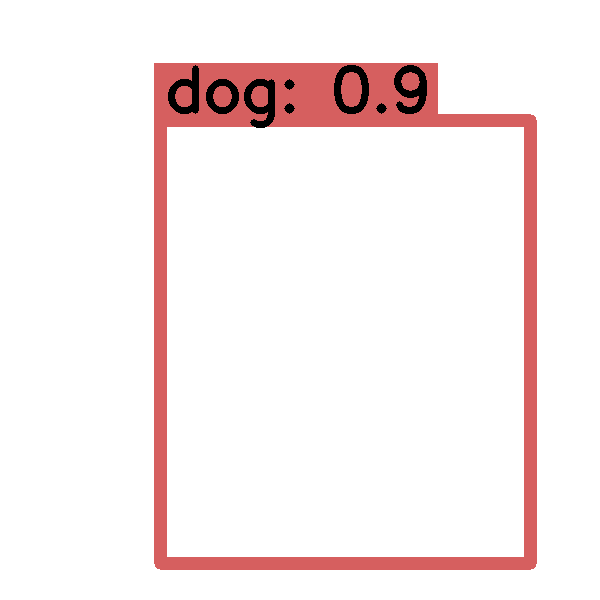} &
        \includegraphics[width=.16\linewidth]{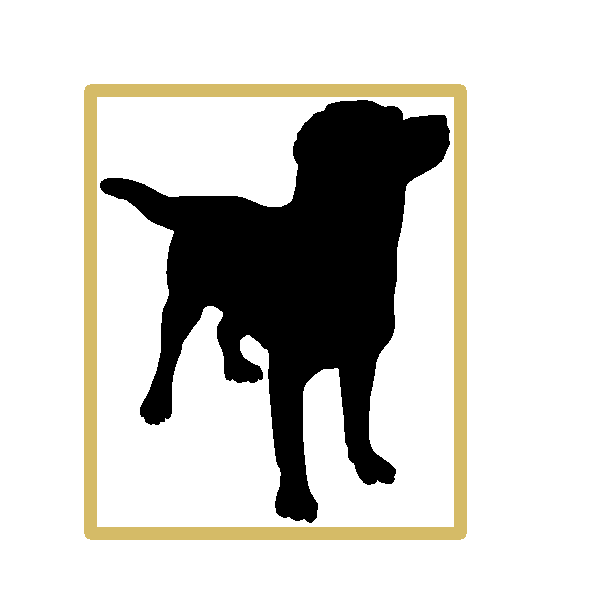} \\
        \includegraphics[width=.16\linewidth]{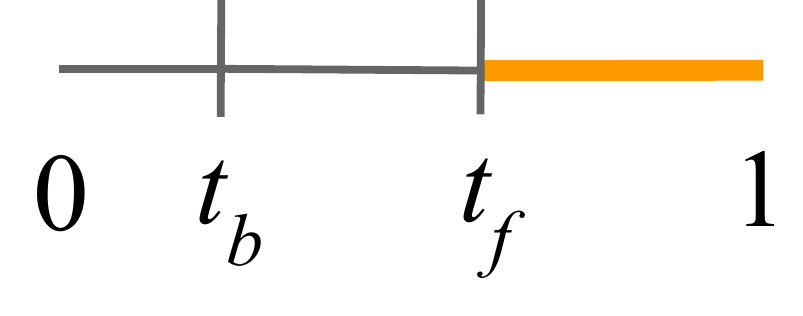}
        &\includegraphics[width=.16\linewidth]{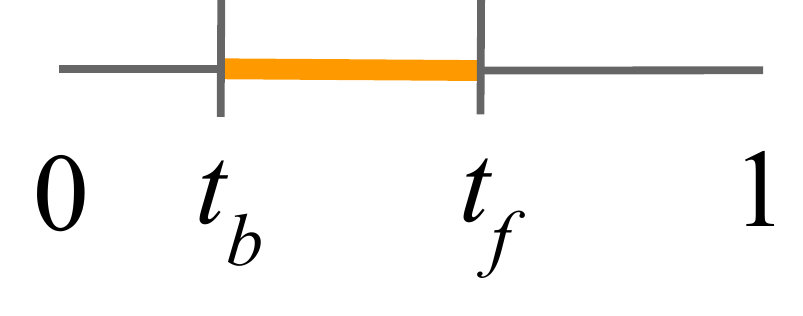}
        & \includegraphics[width=.16\linewidth]{figures/error_types/Iou_localization.pdf}
        & \includegraphics[width=.16\linewidth]{figures/error_types/Iou_classification.pdf}
        & \includegraphics[width=.16\linewidth]{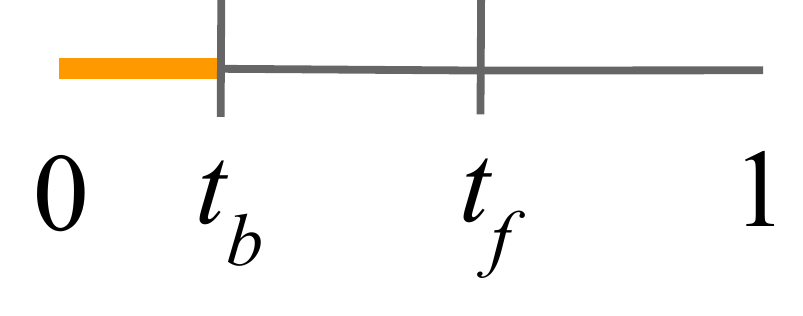}
        & \includegraphics[width=.08\linewidth]{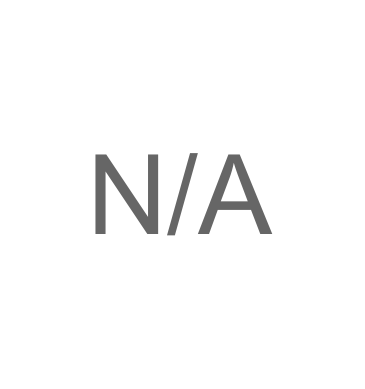}
    \end{tabular}
    \end{center}
    
    \caption{\textbf{Error Type Definitions.} We define 6 error types, illustrated in the \textit{top row}, where box colors are defined as: \crule[false_color]{.25cm}{.25cm} $=$ false positive detection;
    \crule[gt_color]{.25cm}{.25cm}
     $=$ ground truth; \crule[true_color]{.25cm}{.25cm} $=$ true positive detection. The IoU with ground truth for each error type is indicated by an orange highlight and shown in the \textit{bottom row}.
    }
    \label{fig:2_error_type_definition}
\end{figure}
%

%% file: figures/4_progressive_error.tex
\begin{figure}[t]
    \centering

    \subfloat[Default COCO eval style error curves.]{{\includegraphics[width=4.5cm]{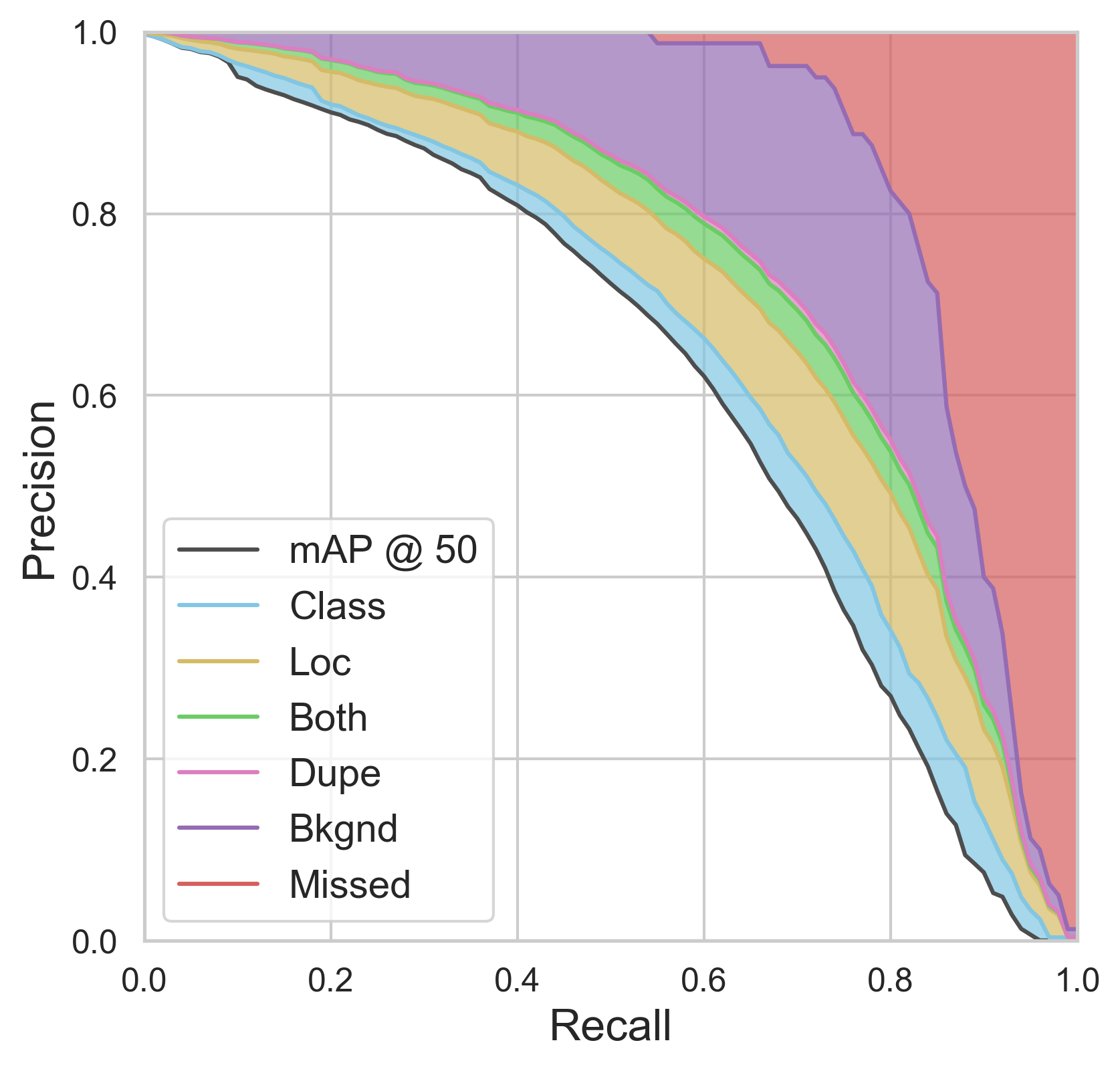} } \label{subfig:4_progressive_original} }%
    \qquad
    \subfloat[Swapping the order of errors changes magnitudes drastically.]{{\includegraphics[width=4.5cm]{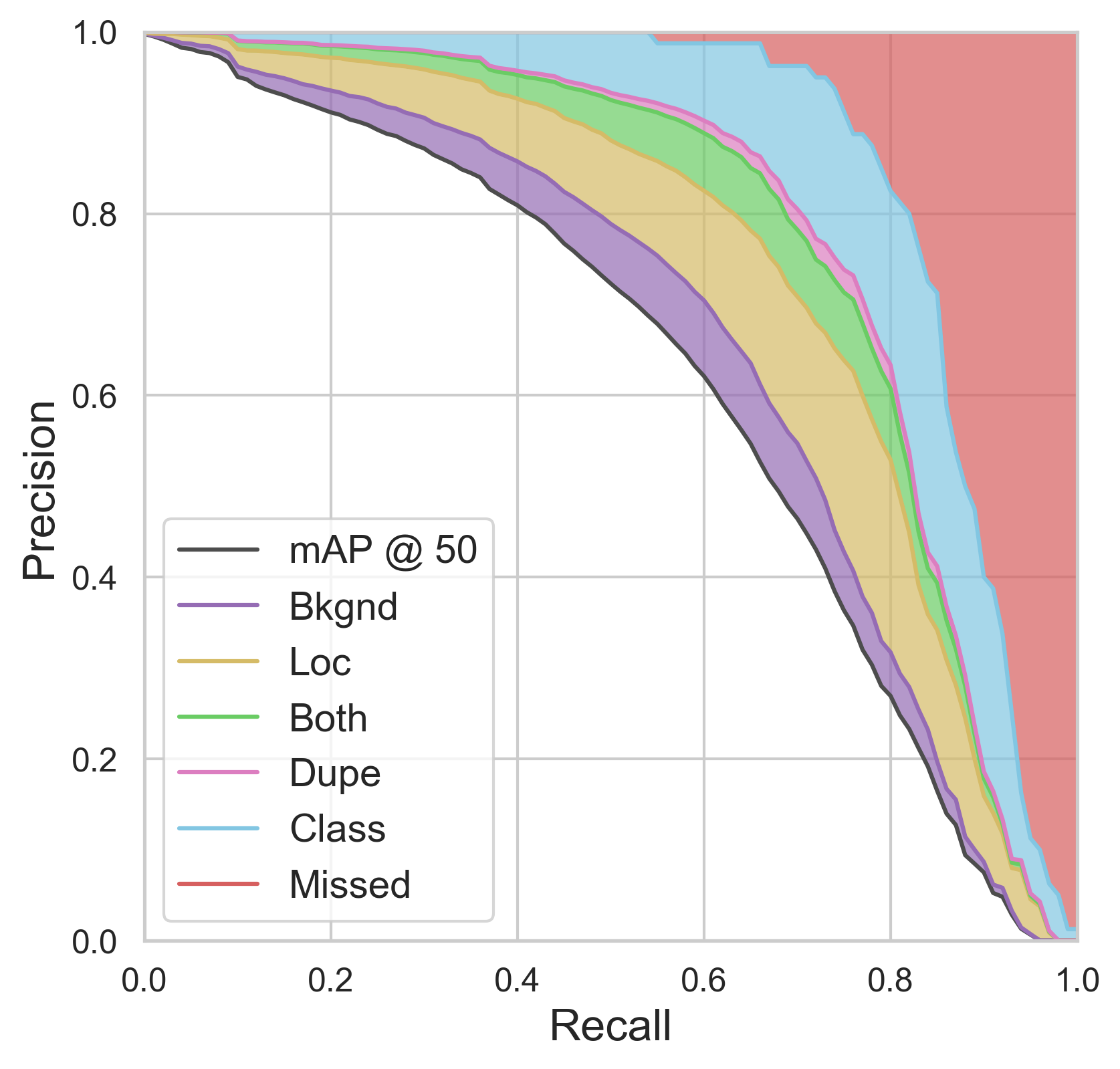} \label{subfig:4_progressive_swapped} }}
    
    \caption{\textbf{The problem with computing errors progressively.} The COCO eval {\tt analyze} function \cite{COCOtoolkit} computes errors progressively, which we show for Mask R-CNN \cite{he2017maskrcnn} detections on $mAP_{50}$. On the right, we swap the order of applying the classification and background oracles. The quantity of each error remains the same, but the perceived contribution from background error (purple region) significantly decreases, while it increases for all other errors. Because COCO computes background error second to last, this instills a belief that it's more important than other errors, which does not reflect reality (see Sec.~\ref{sec:progressive_error}).}
    
    \label{fig:4_progressive_errors}
    \vspace{-.5cm}
\end{figure}

%% file: sections/3_analysis.tex
\section{Analysis}

\input{figures/5_coco_main_errors}

In this section we demonstrate the generality and usefulness of our analysis toolbox by providing detailed analysis across various object detection and instance segmentation models and across different data and annotation sets. We also compare errors based on general qualities of the ground truth, such as object size, and find a number of useful insights. To further explain complicated error cases, we provide more granular analysis into certain error types. 
All modes of analysis used in this paper are available in our toolkit. 

\paragraph{Models} We choose various object detectors and instance segmenters based on their ubiquity and/or unique qualities which allows us to study the performance trade-offs between different approaches and draw several insights. We use Mask R-CNN~\cite{he2017maskrcnn} as our baseline, as many other approaches build on top of the standard R-CNN framework. We additionally include three such models: Hybrid Task Cascades (HTC)~\cite{chen2019htc}, TridentNet~\cite{li2019tridentnet}, and Mask Scoring R-CNN (MS-RCNN)~\cite{he2017maskrcnn}. We include HTC due to its strong performance, being the 2018 COCO challenge winner. We include TridentNet~\cite{li2019tridentnet} as it specifically focuses on increasing scale-invariance. Finally, we include MS R-CNN as a method which specifically focuses on fixing calibration based error. Distinct from the two-stage R-CNN style approaches, we also include three single-stage approaches, YOLACT/YOLACT++~\cite{bolya2019yolact,bolya2019yolact++} to represent real-time models, RetinaNet \cite{lin2017retinanet} as a strong anchor-based model, Fully Convolutional One-Stage Object Detection (FCOS)~\cite{tian2019fcos} as a non anchor-based approach. Where available, we use the ResNet101 versions of each model. Exact models are indicated in the \Appendix.

\paragraph{Datasets} We present our core cross-model analysis on MS-COCO~\cite{lin2014coco}, a widely used and active benchmark. In addition, we seek to showcase the power of our toolbox to perform cross-dataset analysis by including three additional datasets: Pascal VOC~\cite{everingham2010pascal} as a relatively simple object detection dataset, Cityscapes~\cite{cordts2016cityscapes} providing high-res, densely annotation images with many small objects, and LVIS~\cite{gupta2019lvis} using the same images at COCO but with a massive diversity of annotated objects with 1200+ mostly-rare class.

    \subsection{Validating Design Choices}
    \label{sec:validating_design}
        
        \input{figures/8_coco_all_models}

    The authors of each new object detector or instance segmenter make design choices they claim to affect their model's performance in different ways. While the goal is almost always to increase overall $mAP$, there remains the question: does the intuitive justification for a design choice hold up? In Fig.~\ref{fig:5_coco_main_errors} we present the distribution of errors for all object detectors and instance segmenters we consider on COCO \cite{lin2014coco}, and in this section we'll analyze the distribution of errors for each detector to see whether our errors line up with the intuitive justifications.
    
    \paragraph{R-CNN Based Methods}
    First, HTC \cite{chen2019htc} makes two main improvements over Mask R-CNN: 1.) it iteratively refines predictions (i.e., a cascade) and passes information between all parts of the model each time, and 2.) it introduces a module specifically for improved detection of foreground examples that look like background. Intuitively, (1) would improve classification and localization significantly, as the prediction and the features used for the prediction are being refined 3 times. And indeed, the classification and localization errors for HTC are the lowest of the models we consider in Fig.~\ref{fig:8_coco_all_models} for both instance segmentation and detection. Then, (2) should have the effect of eliciting higher recall while potentially adding false positives where something in the background was misclassified as an object. And this is exactly what our errors reveal: HTC has the lowest missed GT error while having the highest background error (not counting YOLACT++, whose distribution of errors is quite unique).
    
    Next, TridentNet \cite{li2019tridentnet} attempts to create scale-invariant features by having a separate pipeline for small, medium, and large objects that all share weights. Ideally this would improve classification and localization performance for objects of different scales. Both HTC and TridentNet end up having the same classification and localization performance, so we test this hypothesis further in Sec.~\ref{sec:fine_analysis}. Because HTC and TridentNet make mostly orthogonal design choices, they would likely compliment each other well.
    
    \paragraph{One-Stage Methods}
    RetinaNet \cite{lin2017retinanet} introduces focal loss that down-weights confident examples in order to be able to train on all background anchor boxes (rather than the standard 3 negative to 1 positive ratio). Training on all negatives by itself should cause the model to output fewer background false positives, but at the cost of significantly lower recall (since the detector would be biased toward predicting background). The goal of focal loss then is to train on all negatives without causing extra missed detections. We observe this is successful as RetinaNet has one of the lowest background errors across models in Fig.~\ref{subfig:coco_detection}, while retaining slightly less missed GT error than Mask R-CNN.
    
    \input{figures/6_example_ablation}
    
    Then FCOS \cite{tian2019fcos} serves as a departure from traditional anchor-based models, predicting a newly defined box at each location in the image instead of regressing an existing prior. While the primary motivation for this design choice was simplicity, getting rid of anchor boxes has other tangible benefits. For instance, an anchor-based detector is at the mercy of its priors: if there is no applicable prior for a given object, then the detector is likely to completely miss it. FCOS on the other hand doesn't impose any prior-based restriction on its detections, leading to it having one of the lowest missed detection errors of all the models we consider (Fig.~\ref{subfig:coco_detection}). Note that it also has the highest duplication error because it uses an NMS threshold of 0.6 instead of the usual 0.5.

    \paragraph{Real-Time Methods}
    YOLACT~\cite{bolya2019yolact} is a real-time instance segmentation method that uses a modified version of RetinaNet as its backbone detector without focal loss. YOLACT++~\cite{bolya2019yolact++} iterates on the former and additionally includes mask scoring (discussed in Tab.~\ref{tab:6_example_ablation}). Observing the distribution of errors in Fig.~\ref{fig:5_coco_main_errors}, it appears that design choices  employed to speed up the model result in a completely different distribution of errors w.r.t. RetinaNet. Observing the raw magnitudes in Fig.~\ref{subfig:coco_detection}, this is largely due to YOLACT having much higher localization and missed detection error. However, the story changes when we look at instance segmentation, where it localizes almost as well as Mask R-CNN despite the bad performance of its detector (see \Appendix). This substantiates their claim that YOLACT is more conducive to high quality masks and that its performance is likely limited by a poor detector.

    \paragraph{A Note on Ablations}
    To demonstrate the potential usefulness of this toolkit for isolating error contribution and debugging, we showcase how an ablation over error types instead of only over $mAP$ provides meaningful insights while still being compact. As an example, consider the trend of rescoring a mask's confidence based on its predicted IoU with a ground truth, as in Mask Scoring R-CNN \cite{huang2019msrcnn} and YOLACT++ \cite{bolya2019yolact++}. This modification is intended to increase the score of good quality masks and decrease the score of poor quality masks, which intuitively should result in better localization. In order to justify their claims, the authors of both papers provide qualitative examples where this is the case, but limit quantitative support to the change to an observed increase in $mAP$. Unfortunately, a change in $mAP$ alone does not illuminate the cause of that change, and some ablations may show little change in $mAP$ despite the method working. By adding the error types that were affected by the change to ablation tables (e.g., see Tab.~\ref{tab:6_example_ablation}) we not only provide quantitative evidence for the design choice, but also reveal side effects (such that classification calibration error went up), which were previously hidden by the raw increase in $mAP$.
    
   \subsection{Comparing Object Attributes for Fine Analysis}
    \label{sec:fine_analysis}
    
        \input{figures/10_htc_vs_tridentnet_scale}

    In order to compare performance across object attributes such as scale or aspect ratio, the typical approach is to compute $mAP$ on a subset of the detections and ground truth that have the specified attributes (with effective comparison requiring normalized $mAP$ \cite{hoiem2012diagnosing}). While we offer this mode of analysis in our toolkit, this doesn't describe the effect of that attribute on overall performance, just how well a model performs on that attribute. Thus, we propose an additional approach based on the tools we defined earlier for summarizing error's affect on overall performance: simply fix errors and observe $\Delta mAP$ as before, but only those whose associated prediction or ground truth have the desired attribute.
    
    \textbf{Comparing Across Scale}
    As an example of using this approach across different scales of objects, we return to the case of TridentNet vs. HTC discussed in Sec.~\ref{sec:validating_design}. Both models have the same classification and localization error and we would like to understand where the difference, if any, lies. Since TridentNet focuses specifically on scale-invariance, we turn our attention to performance across scales.
    We define objects with pixel areas of between 0 and $16^2$ as extra small (XS), $16^2$ to $32^2$ as small (S), $32^2$ to $96^2$ as medium (M), $96^2$ to $288^2$ as (L), and $288^2$ and above as extra large (XL). In Fig.~\ref{fig:10_htc_vs_tridentnet_scale} we apply our approach across HTC and TridentNet (with Mask R-CNN detections included for reference). This comparison reveals that TridentNet localizes and classifies medium sized objects better than HTC, while HTC is better at large objects. This could potentially be why the authors of TridentNet find that they can achieve nearly the same performance by only evaluating their branch for medium sized objects \cite{li2019tridentnet}.
    Other comparisons between subsets of detections such as across aspect ratios, anchor boxes, FPN layers, etc. are possible with the same approach.

    \subsection{Comparing Performance Across Datasets}
    \label{sec:dataset_comparison}
    
        \input{figures/9_maskrcnn_all_datasets}
    
    Our toolkit is dataset agnostic, allowing us to compare the same model across several datasets, as in Fig.~\ref{fig:9_maskrcnn_all_datasets}, where we compare Mask R-CNN (Faster R-CNN for Pascal) across Pascal VOC~\cite{everingham2010pascal}, COCO~\cite{lin2014coco}, Cityscapes~\cite{cordts2016cityscapes}, and LVIS~\cite{gupta2019lvis}. 
    
    In this comparison, the first immediately clear pattern is that Background error decreases both in overall prevalence (pie charts) and absolute magnitude (bar charts) with increasing density of annotations. Faster R-CNN on Pascal is dominated by background error, but of reduced concern on COCO. Both LVIS and Cityscapes, which are very densely annotated, have almost no background error at all. This potentially indicates that much of the background error in Pascal and COCO are simply due to unannotated objects (see Sec.~\ref{sec:bad_annotations}).
    
    As expected, missed ground truths is a large issue for densely annotated datasets like Cityscapes. The core challenge on Cityscapes is the presence of many small objects, which are well known to be difficult to detect with modern algorithms. On the other hand, LVIS's challenge is missing ground truth by failing to classify detections properly, likely due to its enormous number of classes and its very long tail.\footnote{Note that this conclusion has changed from the original version of the paper, due an oversight in how ignored detections are handled in the toolkit (see the \Appendix{} for details).} The trend of densely annotated datasets being biased toward missing ground truth is also represented in the false positive and false negative error distributions (vertical bars). Overall, Pascal is heavily biased toward false positives, COCO is mixed, and LVIS and Cityscapes are both biased toward false negatives. Note that LVIS is biased toward false negatives even though classification error dominates because most of the false positives are being ignored in AP calculations due to how LVIS is evaluated, meaning that on LVIS, you aren't penalized nearly as much for false positives.
    
    On COCO, Mask R-CNN has a harder time localizing masks (Fig.~\ref{subfig:maskrcnn_all_datasets_segmentation}) than boxes (Fig.~\ref{subfig:maskrcnn_all_datasets_detection}), but the opposite is true for LVIS, possibly because of its higher quality masks, which are verified with expert studies \cite{gupta2019lvis}. Again, this potentially indicates that a lot of the error in instance segmentation may be derived by mis-annotations.

    \subsection{Unavoidable Errors}
    \label{sec:bad_annotations}
    
        \input{figures/3_bad_annotations}
    
    We find in Sec.~\ref{sec:dataset_comparison} that a lot of the background and localization error may simply be due to mis- or unannotated ground truth. Examining the top errors more closely, we find that indeed (at least in COCO), many of the most confident errors are actually misannotated or ambiguously misannotated ground truth (see Fig.~\ref{fig:3_bad_annotations}). For instance, 30 of the top 100 most confident localization errors in Mask R-CNN detections are due to bad annotations, while the number soars to 50 out of 100 for background error. These misannotations are simple mistakes like making the box too big or forgetting to mark a box as a crowd annotation. More examples are ambiguous: should a mannequin or action figure be annotated as a person? Should a sculpture of a cat be annotated as a cat? Should a reflection of an object be annotated as that object? Highly confident mistakes result in large changes in overall $mAP$, so misannotated ground truth considerably lower the maximum $mAP$ a reasonable model can achieve.
    
    This begs the question, what is the upper bound for mAP on these datasets? Existing analyses into the potential upper bound in object detection such as \cite{borji2019empirical} don't seem to account for the rampant number of mislabeled examples. The state-of-the-art on the COCO challenge are slowly stagnating, so perhaps we are nearing the ``reasonable'' upper bound for these detectors. We leave this for future work to analyze.

%% file: figures/5_coco_main_errors.tex
\begin{figure}[!tbp]
    \centering

    \includegraphics[width=\textwidth]{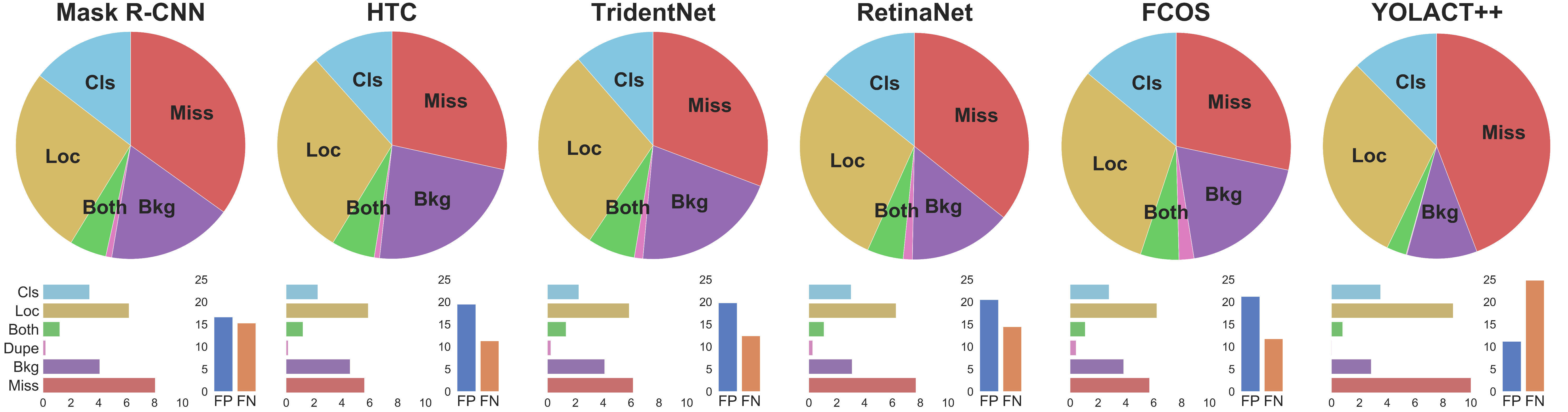} 
    
    \caption{\textbf{Summary of errors on COCO Detection.} Our model specific error analysis applied to various object detectors on COCO. The pie chart shows the relative contribution of each error, while the bar plots show their absolute contribution. For instance segmentation results, see the \Appendix.
    }
    
    \label{fig:5_coco_main_errors}
\end{figure}

%% file: figures/8_coco_all_models.tex
\begin{figure}[!t]
    \centering
    
    \subfloat[Detection Performance.]{
        \includegraphics[width=0.5\textwidth]{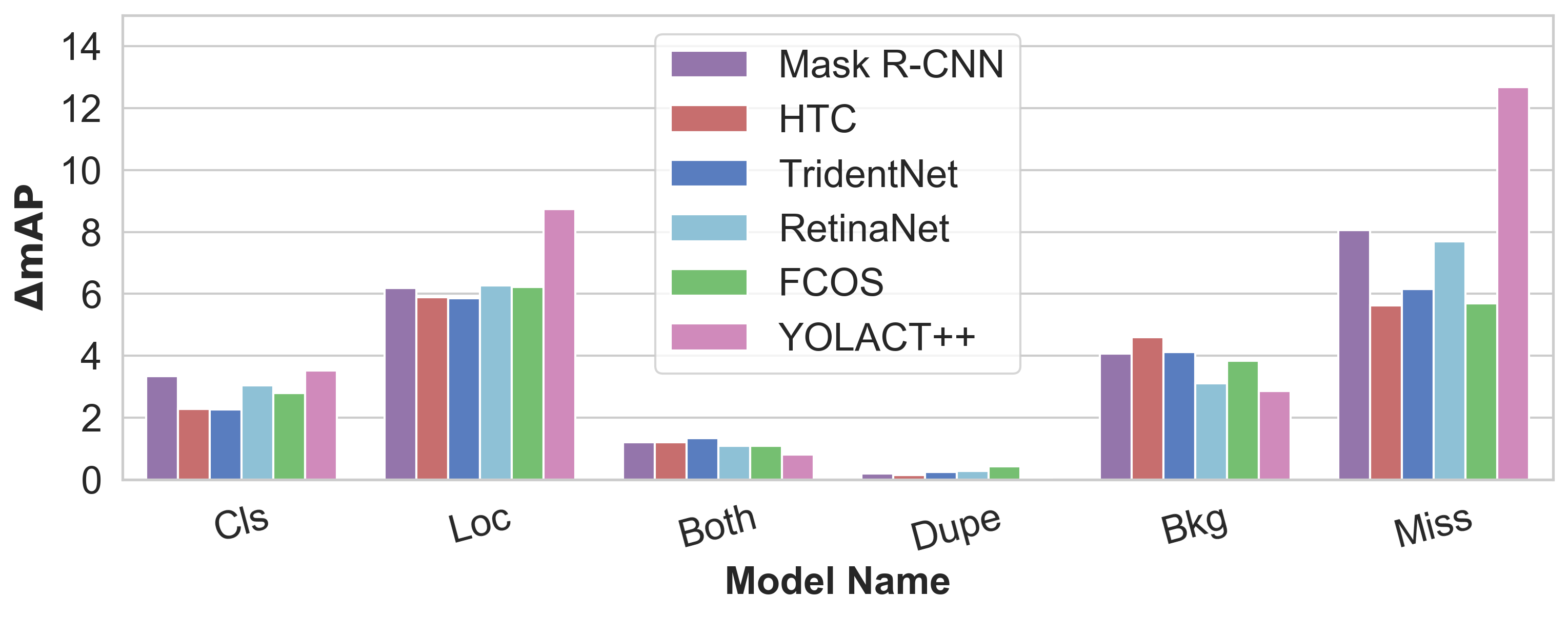}
        \label{subfig:coco_detection}
    }%
    \subfloat[Instance Segmentation Performance.]{
        \includegraphics[width=0.5\textwidth]{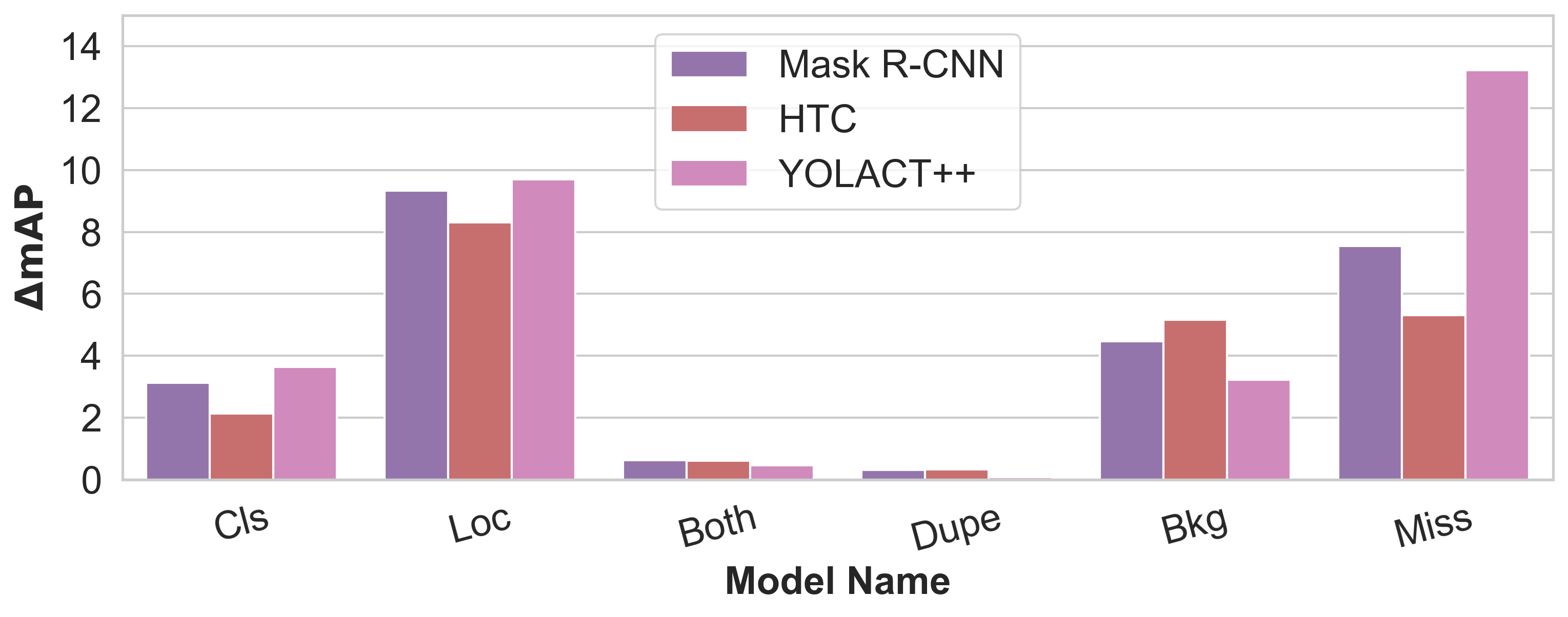}
        \label{subfig:coco_segmentation} 
    }
    
    \caption{\textbf{Comparison across models on COCO.} Weight of each error type compared across models. This has the same data as Fig.~\ref{fig:5_coco_main_errors}. }
    \vspace{-.5cm}
    \label{fig:8_coco_all_models}
\end{figure}

%% file: figures/6_example_ablation.tex
\begin{table}[!tb]
    \centering
    
    \newcommand{\uparr}{\rotatebox[origin=c]{90}{\ding{213}}}
    \newcommand{\downarr}{\rotatebox[origin=c]{270}{\ding{213}}}
    
    \newcommand{\dmAP}[1]{\text{E}_\texttt{#1}{\text{\downarr}}}
    \newcommand{\spacer}{\phantom{aa}}
    \newcommand{\halfspacer}{\phantom{a}}
    
    \caption{{\bf Mask Rescoring.} An ablation of MS-RCNN~\cite{he2017maskrcnn} and YOLACT++~\cite{bolya2019yolact++} mask performance using the errors defined in this paper. $\Delta mAP_{50}$ is denoted as E for brevity, and only errors that changed are included. Mask scoring better calibrates localization, leading to decrease in localization error. However, by scoring based on localization, the calibration of other error types suffer. Note that this information is impossible to glean from the change in $\text{AP}_{50}$ alone. }
    
    \begin{smalltable}{l r r r rrrr rr rr} \toprule 
                  Method && $\text{AP}_{50}{\text{\uparr}}$ && $\dmAP{cls}$ & $\dmAP{loc}$ & $\dmAP{bkg}$ & $\dmAP{miss}$ &&& $\dmAP{FP}$ & $\dmAP{FN}$ \\
    \midrule
 Mask R-CNN (R-101-FPN) &&             58.1 &  &          3.1 &          9.3 &        4.5 &           7.5 &&&        15.9 &        17.8 \\
  $\quad$+ Mask Scoring & &            58.3 &  &          3.6 &          7.8 &        5.1 &           7.8 &&&        15.9 &        18.1 \\
    Improvement        &  &         $+0.2$ &  &       $+0.4$ &       $\mathbf{-1.5}$ &     $+0.7$ & $+0.3$ &&&      $+0.0$ &      $+0.3$ \\
\midrule
     YOLACT++ (R-50-FPN) &\spacer&             51.8 &\spacer&          3.3 &         10.4 & 3.2&          13.0 &\halfspacer&\halfspacer&        10.7 &        27.7 \\
      $\quad$+ Mask Scoring &&             52.3 &&          3.6 &          9.7 & 3.2&          13.2 &&&        10.1 &        28.2 \\
      Improvement           &&           $+0.5$ &  &       $+0.3$ &       $\mathbf{-0.7}$ & $+0.0$ &        $+0.2$ &&&      $-0.5$ &      $+0.6$ \\
\bottomrule
\end{smalltable}
    
    \label{tab:6_example_ablation}
\end{table}

%% file: figures/10_htc_vs_tridentnet_scale.tex
\begin{figure}[tbp]
    \centering
    
    \subfloat[Classification Error]{\includegraphics[width=0.49\textwidth]{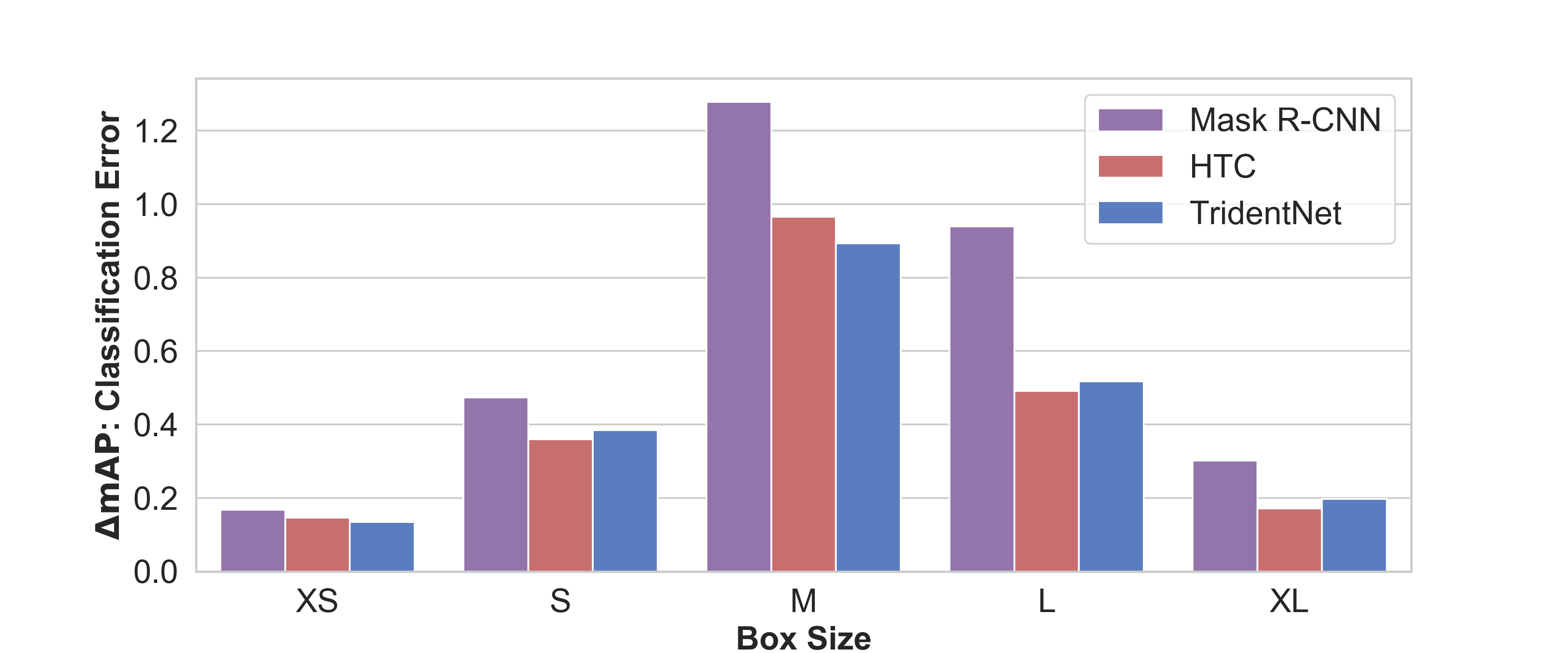}}
    \subfloat[Localization Error]{\includegraphics[width=0.49\textwidth]{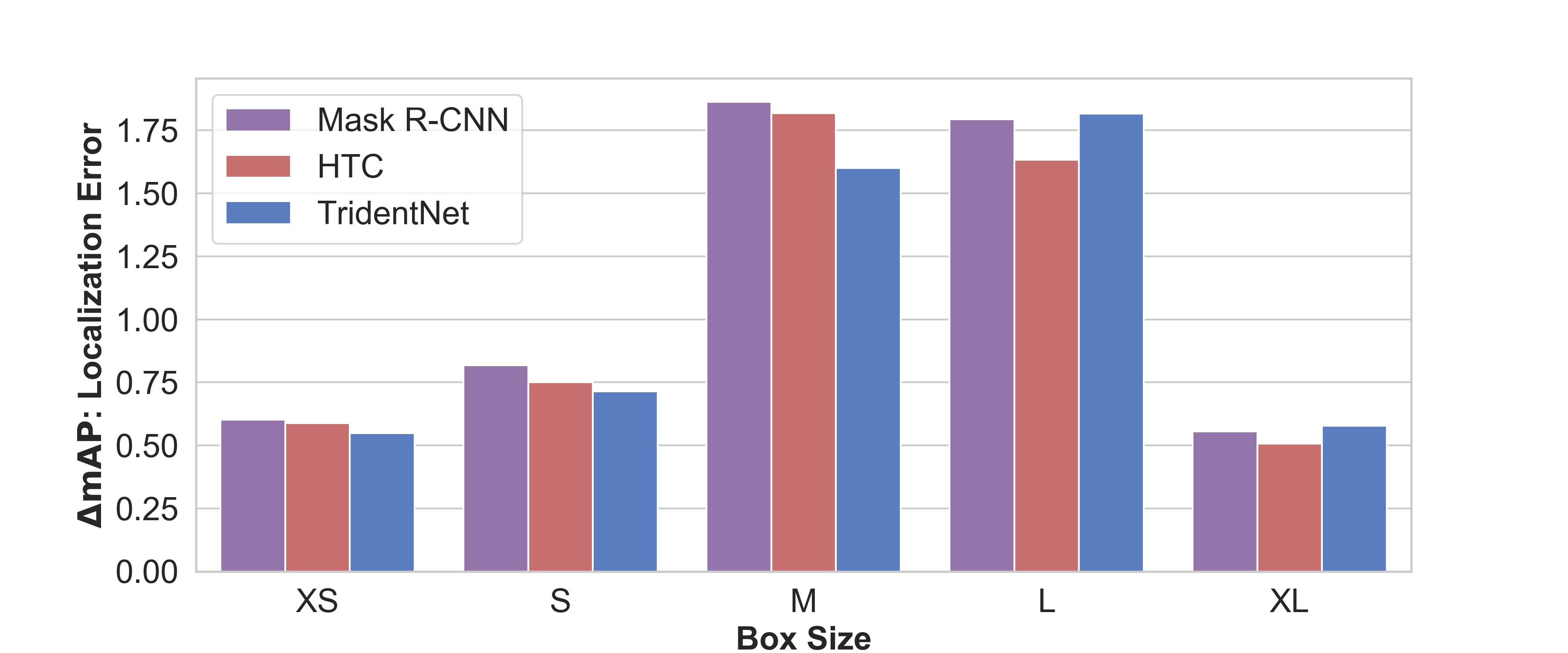}}
    \caption{\textbf{Comparison of Scales between HTC and TridentNet.} Both HTC and TridentNet have the same classification and localization error on COCO detection. Using fine analysis, we can isolate the cause of these errors further. }
    \vspace{-.5cm}
    \label{fig:10_htc_vs_tridentnet_scale}
\end{figure}

%% file: figures/9_maskrcnn_all_datasets.tex
\begin{figure}[!tb]
    \centering
    
    \begin{tabular}{c:c}
    \subfloat[Object Detection]{
        \includegraphics[width=0.57\textwidth]{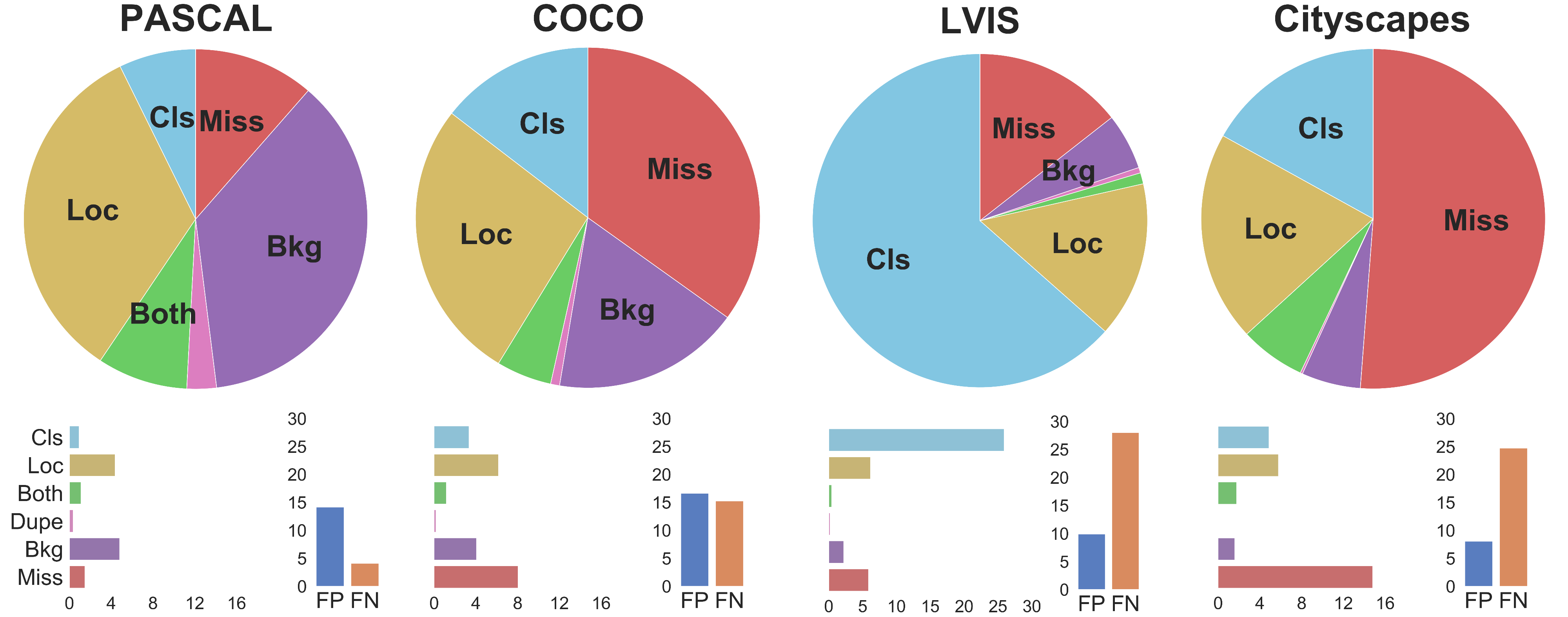}
        \label{subfig:maskrcnn_all_datasets_detection}
    }%
    &
    \subfloat[Instance Segmentation]{
        \includegraphics[width=0.43\textwidth]{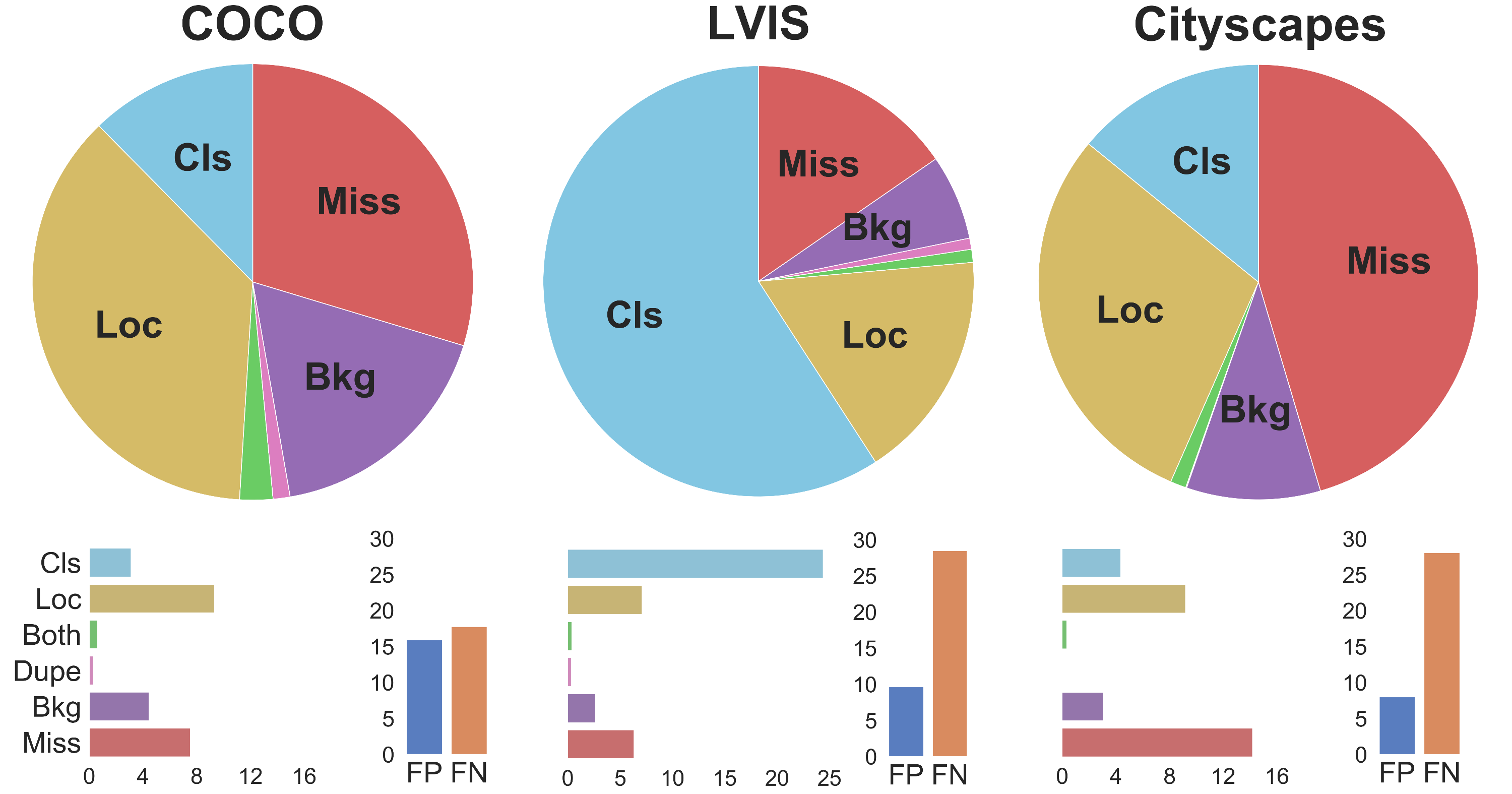}
        \label{subfig:maskrcnn_all_datasets_segmentation}
    }
    \end{tabular}
    
    \caption{\textbf{Performance of Mask R-CNN Across Datasets.} Because our toolkit is dataset agnostic, we can fix a detection architecture and compare performance across datasets to gain valuable insights into properties of each dataset. }
    \vspace{-.5cm}
    \label{fig:9_maskrcnn_all_datasets}
\end{figure}

%% file: figures/3_bad_annotations.tex
\begin{figure}[t]
    \centering


    
    \includegraphics[width=.9\textwidth]{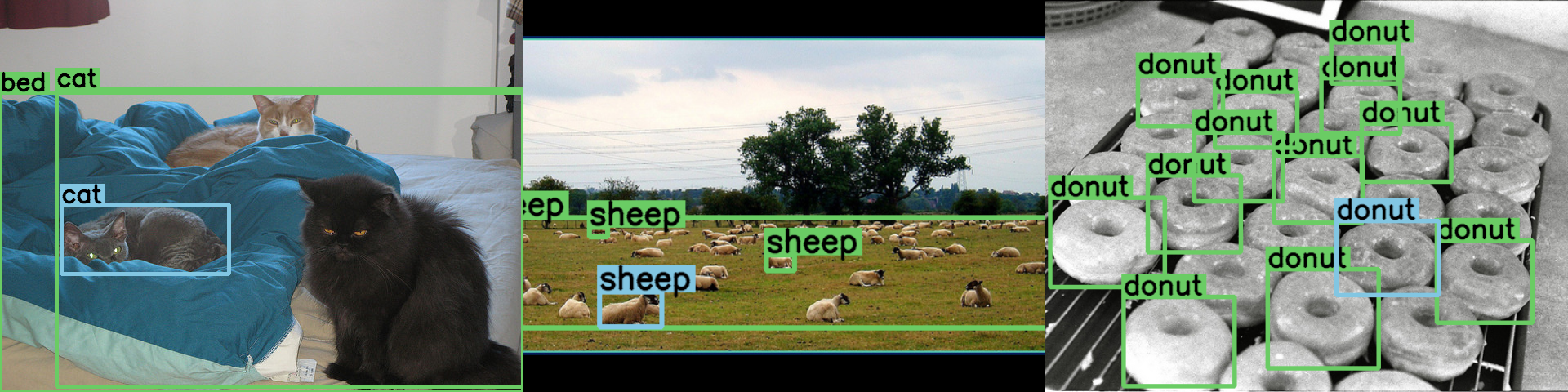}
    
    \caption{\textbf{Examples of Poor Annotations.} In modern detectors, highly confident detections classified as both mislocalized and misclassified or background errors are likely to be mislabeled examples on COCO. In the first two images, the ground truth should have been labeled as crowds. In the third, some of the donuts simply weren't labeled. \crule[true_color]{.25cm}{.25cm} $=$ ground truth, \crule[pred_color]{.25cm}{.25cm} $=$ predictions.}
    
    \label{fig:3_bad_annotations}
\end{figure}

%% file: sections/4_conclusion.tex
\section{Conclusion}

In this work, we define meaningful error types and a way of tying these error types to overall performance such that it minimizes any confounding variables. We then apply the resulting framework to evaluate design decisions, compare performance on object attributes, and reveal properties of several datasets, including the prevalence of misannotated ground truth in COCO. We hope that our toolkit can not only serve as method to isolate and improve on errors in detection, but also lead to more interpretability in design decisions and clearer descriptions of the strengths and weaknesses of a model.

%% file: sections/5_appendix.tex
\appendix

\section{Additional Discussion}

    There are some minor details left out of the main paper due to space constraints. We discuss those details here.

    \subsection{Oddities of $\Delta AP$}
    
        While $\Delta AP_a$ correctly weights the importance of error type $a$, it has some potentially unintuitive properties that we list here.
        
        First, consider Tab.~2 in the main paper. It would be nice if the improvement (or negative improvement) for each error type when summed equaled the overall improvement in $AP$. For instance, take the improvement row for MS R-CNN. The improvement in $AP_{50}$ is $+0.2$, while the sum of changes in main errors ($-(0.4-1.5+0.7+0.3)$) is $+0.1$. Using the special errors (FP and FN, $-(0.0+0.3)$) would even predict worse $AP_{50}$: $-0.3$. That is, in general
        \begin{equation}
            \Delta AP \neq - \sum_{o \in \mathcal{O}}{\Delta E_o}
        \end{equation}
        This isn't a huge issue, as the relative magnitudes of each error type can still be compared. However, it's something that needs to be kept in mind while analyzing performance.
        
        Second, and a related issue is that summing the $\Delta AP_a$ for each error type $a$ does not result in $100 - AP$. For instance, consider the special error types (FP and FN), which should account for all the error in the model. If we use the numbers from the same Tab.~2 for Mask R-CNN (first row), adding the $AP_{50}$ with $\Delta AP_{FP}$ and $\Delta AP_{FN}$ ($58.1+15.9+17.8$) yields 91.8, not 100. Similarly, for YOLACT++ we have ($51.8+10.7+27.7$) 90.2, which is again not 100. More concretely, for $\mathcal{O} = \{o_1, \ldots, o_n\}$ this means in general
        \begin{equation}
            AP + \Delta AP_{o_1} + \ldots + \Delta AP_{o_n} \neq 100
        \label{sup:eq:apsum}
        \end{equation}
        
        This is a direct result of not computing errors progressively (Fig.~2 in the main paper), where the errors sum to 100, and in fact is an odd property of AP explained in Sec.~2.3 of the main paper: fixing multiple errors at once gives a bigger boost in $mAP$ than fixing each error on their own.
        
        Both of these issues have an underlying cause that we can see if we write out the same expression as in Eq.~\ref{sup:eq:apsum} but with progressive error:
        \begin{equation}
            AP + \Delta AP_{o_1, \ldots, o_n} = 100
        \label{sup:eq:apsumprog}
        \end{equation}
        which begs the question, how can we relate $\Delta AP_a + \Delta AP_b$ to $\Delta AP_{a,b}$? It turns out that they differ by $(\Delta AP_a - \Delta AP_{a \mid b})$.
        
        To show this, we first split each term into its definition:
        \begin{equation}
            \Delta AP_{a, b} = AP_{a, b} - AP \qquad \Delta AP_a + \Delta AP_b = AP_a + AP_b - 2AP
        \label{eq:delAP_linearity1}
        \end{equation}
        Then we rearrange the terms for the left equation to get it in terms of AP:
        \begin{equation}
            AP = AP_{a, b} - \Delta AP_{a, b}
        \label{eq:delAP_linearity2}
        \end{equation}
        Then, substitute 1 AP into the right equation in Eq.~\ref{eq:delAP_linearity1} to get
        \begin{equation}
            \Delta AP_a + \Delta AP_b = AP_a + AP_b - AP - AP_{a, b} + \Delta AP_{a, b}
        \label{eq:delAP_linearity3}
        \end{equation}
        We can then group $AP_a - AP$ and $-(AP_{a,b} - AP_b)$ and substitute them with the definitions for $\Delta AP_a$ and $-\Delta AP_{a \mid b}$ respectively (if collecting the terms a different way we could swap $a$ and $b$ here). This leaves us with the following:
        \begin{equation}
            \Delta AP_a + \Delta AP_b = \Delta AP_{a, b} + (\Delta AP_a - \Delta AP_{a \mid b})
        \label{eq:delAP_linearity4_final}
        \end{equation}
        Since the $\Delta AP_{a \mid b} > \Delta AP_a$ in most cases (following the reasoning given in Sec.~2.3), this means $\Delta AP_{a, b} > \Delta AP_a + \Delta AP_b$ in most cases.
        
        Oddities like this are why such great care needs to be taken when working with $AP$, since the properties it has are not intuitive.
        
        \begin{table}[t]
            \newcommand{\dmAP}[1]{\text{E}_\texttt{#1}}
            \newcommand{\spacer}{\phantom{aa}}
            \newcommand{\halfspacer}{\phantom{a}}
            
            \centering
            \caption{{\bf Errors over thresholds.} Evaluating the error types at different foreground IoU thresholds ($t_f$) using Mask R-CNN detections on COCO. }
            \begin{smalltable}{c c r c rrrrrr c rr} \toprule
$t_f$&\spacer&AP&\spacer&$\dmAP{cls}$&$\dmAP{loc}$&$\dmAP{both}$&$\dmAP{dupe}$&$\dmAP{bkg}$&$\dmAP{miss}$&\spacer&$\dmAP{FP}$&$\dmAP{FN}$\\\midrule
$0.5$&&61.7&&3.3&6.2&1.2&0.2&4.1&7.0&&16.6&15.3\\
$0.6$&&57.1&&2.7&10.6&1.2&0.0&3.5&7.3&&16.5&18.3\\
$0.7$&&49.7&&2.1&18.1&0.9&0.0&2.7&7.0&&15.0&23.9\\
$0.8$&&36.1&&1.3&31.6&0.6&0.0&1.4&6.9&&12.9&32.1\\
$0.9$&&12.0&&0.2&55.1&0.1&0.0&0.3&4.9&&9.7&33.2\\
            \bottomrule
            \end{smalltable}\\
            \label{sup:tab:iou_ablation}
        \end{table}

    \subsection{$\text{AP}^{0.5:0.95}$}

        The primary metric used in the COCO and CityScapes challenges is $AP^{0.5:0.95}$, or the average of $mAP$ across 10 IoU thresholds starting from 0.5 to 0.95 with increments of 0.05. All our analysis in our main paper is done with an IoU threshold ($t_f$) of 0.5, but it's worth looking at higher thresholds because of this metric.
        
\begin{figure}[t]
            \centering
            
            \subfloat[Main Error Types]{{\includegraphics[width=0.98\textwidth]{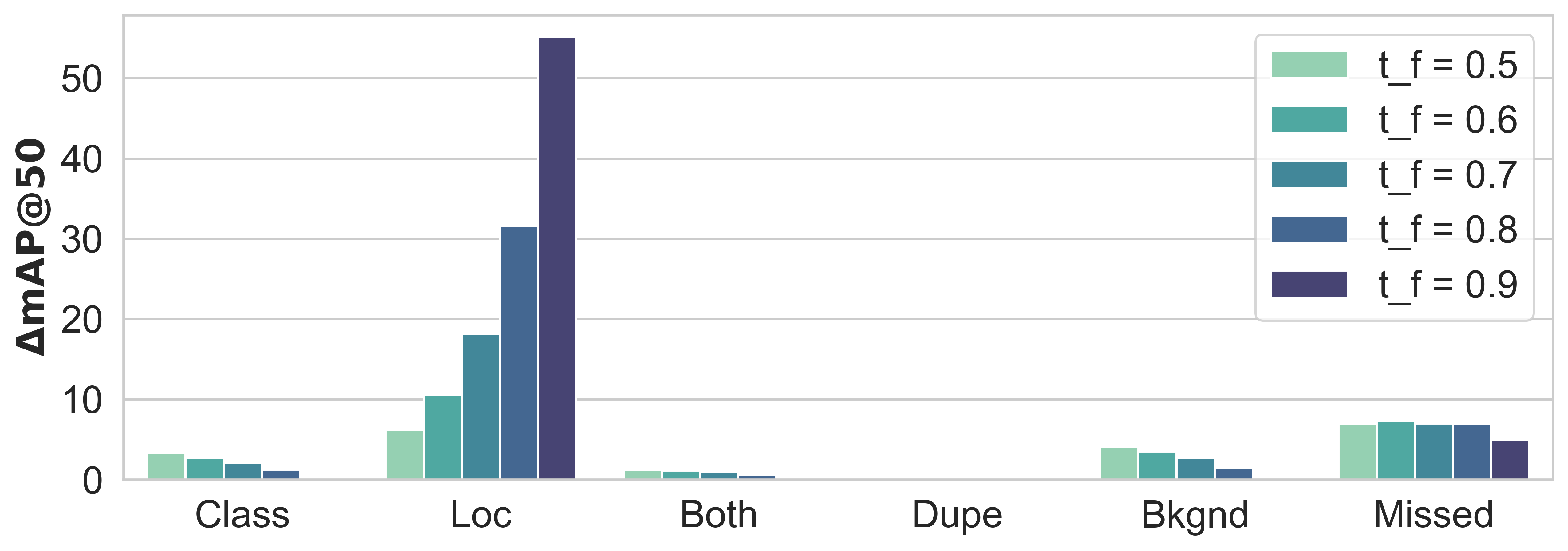} \label{sup:subfig:iou_ablation_main}}}\\
            \subfloat[Special Error Types]{{\includegraphics[width=0.98\textwidth]{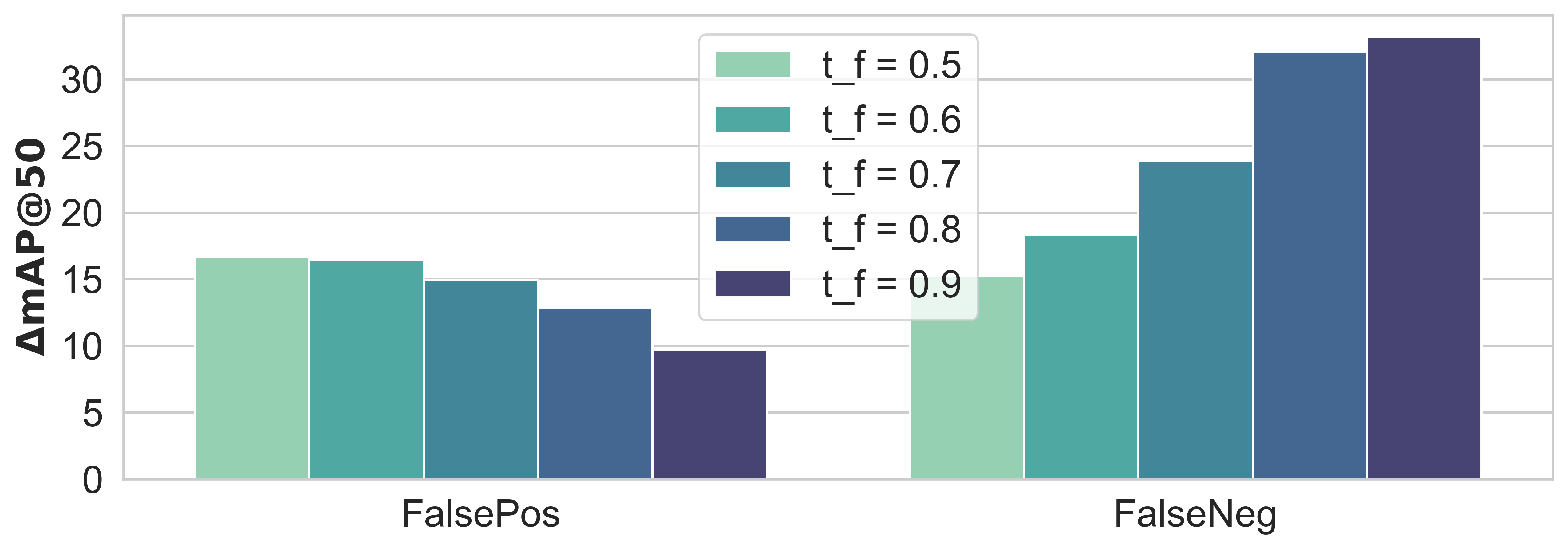} \label{sup:subfig:iou_ablation_special}}}
            
            \caption{{\bf Errors over thresholds.} The values for these plots are reproduced in Tab.~\ref{sup:tab:iou_ablation}. This is using Mask R-CNN detections on COCO.}
            \label{sup:fig:iou_ablation}
        \end{figure}

        In Tab.~\ref{sup:tab:iou_ablation} and Fig.~\ref{sup:fig:iou_ablation} we evaluate the error types over the IoU thresholds 0.5, 0.6, 0.7, 0.8, and 0.9 using Mask R-CNN detections on COCO. As expected, the error type that responds most strongly to higher IoU thresholds is localization error (Fig.~\ref{sup:subfig:iou_ablation_main}), since increasing the threshold just makes it harder to localize to the ground truth. We also see that false negatives start mattering more than false positives at higher IoU thresholds (Fig.~\ref{sup:subfig:iou_ablation_special}).
        
        Thus, COCO and CityScape's average over IoU threshold metric is biased heavily toward localization errors and to a lesser extent false negatives. This is why Mask Scoring R-CNN, which rescores its masks in a way that better calibrates localization at the expense of other error types (see Tab.~2 in the main paper), is so effective. Their approach makes no significant difference at $AP_{50}$, but at the higher thresholds that are biased more toward localization, they get a huge boost, leading to a big improvement in ${AP}^{0.5:0.95}$. Moreover, many aspects of YOLACT / YOLACT++ are much worse than other methods, which we can see by its detector performance in Fig.~\ref{sup:fig:coco_summary}, but it localizes masks on par with other instance segmenters which give it a boost in performance in ${AP}^{0.5:0.95}$ for instance segmentation. Whether this is a desirable trait for a metric is up to the dataset maintainers, but designers need to take this into consideration when prioritizing areas of improvement.

\section{Implementation Details}

    Here we discuss design choices and implementation details that weren't able to fully explain in the main paper.

    \subsection{Defining the Missed GT Oracle}
    
        \begin{figure}[t]
            \centering
        
            \includegraphics[width=.98\linewidth]{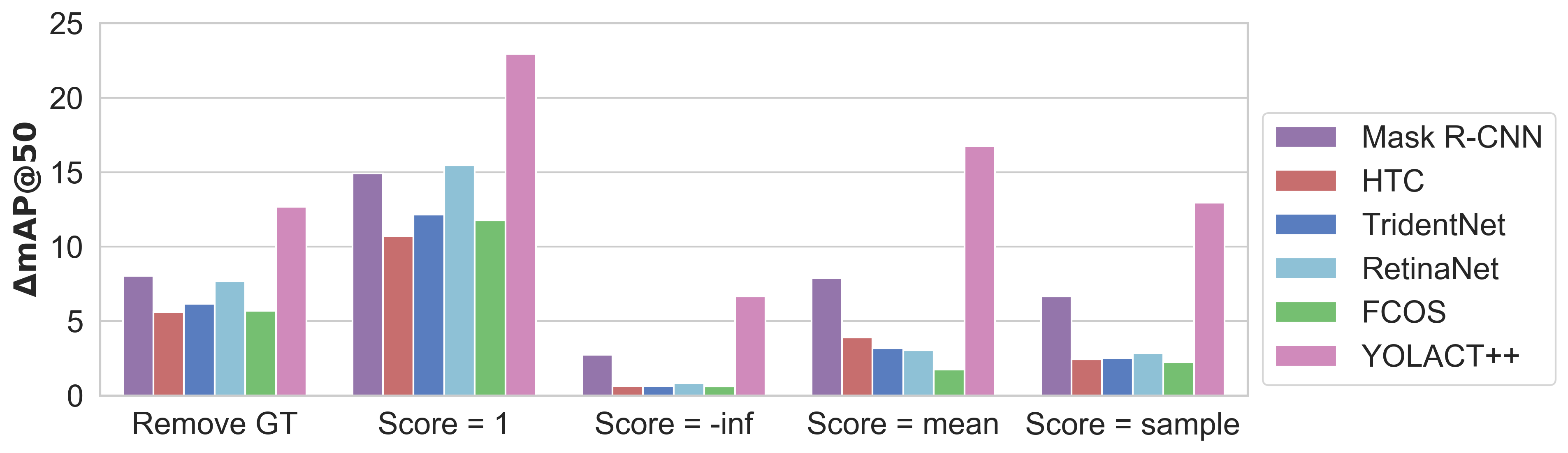}
        
            \caption{\textbf{Possible Definitions for Missed GT.} Defining the oracle for missing GT is difficult, so it's important to choose a good definition. Here we compare ways of choosing the score for a new true positive versus just removing GT (what we use in our main paper). }
            \label{sup:fig:missedgtoracle}
        \end{figure}
    
        As we noted in the main paper, creating a definition for ``fixing'' false negatives is a tricky subject. We outlined two strategies to do so: remove true negatives (i.e., lower $N_{GT}$) or add true positives (i.e., add a detection). We chose the former because the latter required us to choose a score for this new detection. In this section, we elaborate why choosing the right score is difficult and may lead to false conclusions. We do this by evaluating several reasonable techniques for choosing this score.
        
        First, we could set the score to some fixed value. Two obvious choices are 1 (to put all the new detections at beginning of the sorted list) and $-\infty$ (to put all new detections at the end of the list). As evident in Fig.~\ref{sup:fig:missedgtoracle}, setting the score to 1 produces very high values for the missing GT and likely overweights their contribution. In effect, setting the score to 1 assumes that whatever predictions the model add to catch this missed GT will be perfectly calibrated. Since the predictions for the other error types aren't perfectly calibrated, this results in the relative weight for missed GT being too high.
        
        On the other hand, setting the score to $-\infty$ essentially uses the lowest score output by the model. This doesn't just assume the model will have poor calibration for this GT, but it also depends heavily on how many low scoring detections the model produced. In order to boost $AP$, many detectors (HTC, FCOS, TridentNet, RetinaNet) produce a lot of low-scoring detections (since COCO allows 100 detections per image). This results in this version of missed GT being disproportionately small for these models as compared to the rest.
        
        Another, and perhaps more reasonable, method for determining the score samples from the existing predictions' scores. In Fig.~\ref{sup:fig:missedgtoracle} we test setting the score to the mean of the predicted scores and setting the score to that of an existing prediction sampled uniformly at random. However, as we see in Fig.~\ref{sup:fig:missedgtoracle}, both methods produce the same skewed results as simply setting the score to $-\infty$ does (e.g., Mask R-CNN and RetinaNet for removing GT and setting the score to 1 are nearly identical, but wildly different for the other definitions).
        
        In general, we can't trust the distribution of scores given by the detector as accurate, since some detectors like to use all of the available bandwidth of detections (100 per image for COCO) by flooding the predictions with low scoring detections that have some chance of being correct. This produces skewed results when defining the score as anything that depends on these low scoring detections ($-\infty$, mean, sample). Thus, we can't tell what the score for the new prediction should be, leaving us with the only option of defining the missed GT oracle as removing true negatives.
        
    \subsection{Breaking Ties in Error Assignment}
        
        \begin{figure}[t]
            \centering
        
            \includegraphics[width=.4\linewidth]{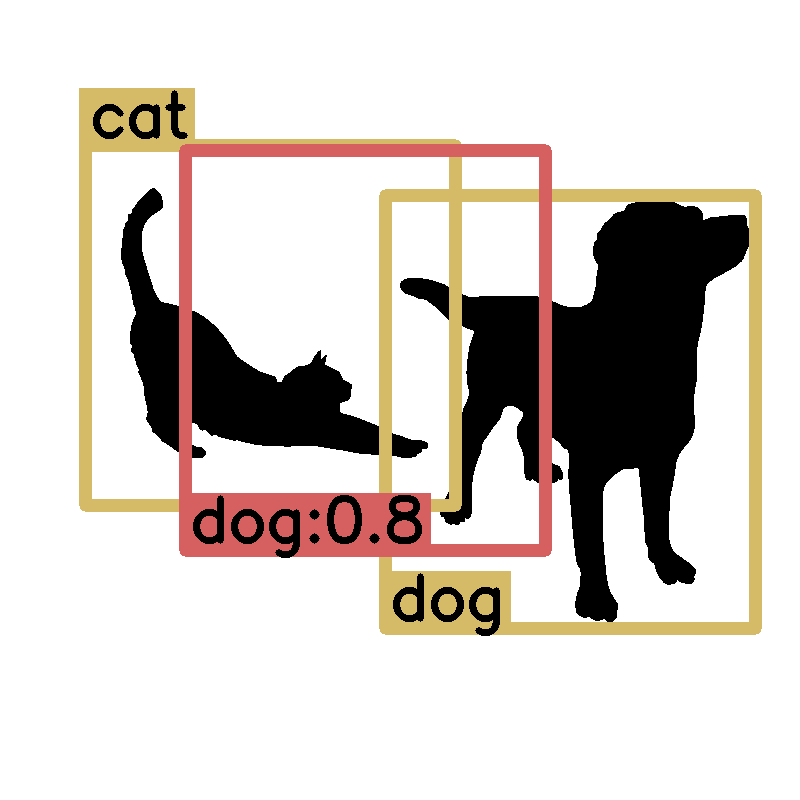}
        
            \caption{\textbf{Ties in Error Assignment.} A possible tie in error assignment is illustrated here (\crule[false_color]{.25cm}{.25cm} $=$ false positive detection; \crule[gt_color]{.25cm}{.25cm} $=$ ground truth). The prediction is a classification error for the cat and a localization error for the dog. We break ties like this by assigning a localization error with the dog. Note that this is different to the both cls+loc error, as these are errors with two entirely separate ground truths. }
            \label{sup:fig:tiedassignment}
        \end{figure}
        
        Some situations can cause a predicted box to have two separate errors with two different ground truth. This is because when computing $IoU_\text{max}$ for classification error, we use GT of a \textit{different} class, while for localization we use the GT of the \textit{same} class. Thus, it's possible for a given prediction to have a localization error w.r.t. one GT and a classification error w.r.t. another GT (as illustrated in Fig.~\ref{sup:fig:tiedassignment}).
        
        While these ties don't happen often ($\sim 0.78\%$ of Mask R-CNN predictions on COCO), it is important to deal with them in a defined way. In the case presented in Fig.~\ref{sup:fig:tiedassignment}, we prioritize localization error over classification error, choosing to trust the classification of the model more than its localization. This choice is largely arbitrary (there are arguments for both) but needs to be made. For all other tie breakers, we follow the order we define the error types.
        
        Note that this tie breaking is not to be confused with computing errors progressively. In our implementation, we first assign an error type to each false positive and false negative, and then only after all positives and negatives are accounted for do we compute $\Delta AP$.
    
    \subsection{Models and Sources Used}

        We used off-the-shelf models for each method tested and didn't train any new models. Some methods directly provided a COCO evaluation JSON file which we could use directly with our toolbox, while others required us to run the code ourselves. For each model, we list the method name, the model description (as describes the relevant weights file), a link to the code we used, and whether or not the method provided a COCO JSON (i.e., whether we didn't need to evaluate the model ourselves or not) in Tab.~\ref{sup:tab:model_sources}. Note that in general we use the Resnet101 version of each model without any bells and whistles. The one exception is YOLACT++, since it uses deformable convolutions while the rest of the models don't, so we use its Resnet50 model to compensate.
        
        \begin{table}[h!]
            \newcommand{\yes}{\ding{52}}
            \newcommand{\no}{\ding{55}}
            
            \newcommand{\detectron}{\href{https://github.com/facebookresearch/Detectron/blob/master/MODEL_ZOO.md}{detectron}}
            
            \centering
            \caption{{\bf Model Sources.} The sources for the models we used in our analysis.}
            \begin{smalltable}{l l l c} \toprule
                Method & Model Description & Implementation & JSON?\\\midrule
                Mask R-CNN & R-101-FPN, 2x (35861858) & \detectron & \yes \\
                MS R-CNN & ResNet-101 FPN & \href{https://github.com/zjhuang22/maskscoring_rcnn}{maskscoring\_rcnn} & \no \\
                HTC & R-101-FPN & \href{https://github.com/open-mmlab/mmdetection/tree/master/configs/htc}{mmdetection} & \no \\
                TridentNet & TridentNet, 1x	(ResNet-101) & \href{https://github.com/TuSimple/simpledet/tree/master/models/tridentnet}{simpledet} & \no \\
                RetinaNet & R-101-FPN, 2x (36768840) & \detectron & \yes \\
                FCOS & FCOS\_R\_101\_FPN\_2x & \href{https://github.com/tianzhi0549/FCOS}{FCOS} & \no \\
                YOLACT++ & Resnet50-FPN & \href{https://github.com/dbolya/yolact}{yolact} & \no \\
            \bottomrule
            \end{smalltable}\\
            \label{sup:tab:model_sources}
        \end{table}

\section{COCO Instance Segmentation Summary}

\begin{figure}[!tbh]
    \centering

    \subfloat[Detection Results]{{
    \includegraphics[width=0.98\textwidth]{figures/images/coco_detection_summary.png} }}\\
    \subfloat[Instance Segmentation Results]{{\includegraphics[width=0.98\textwidth]{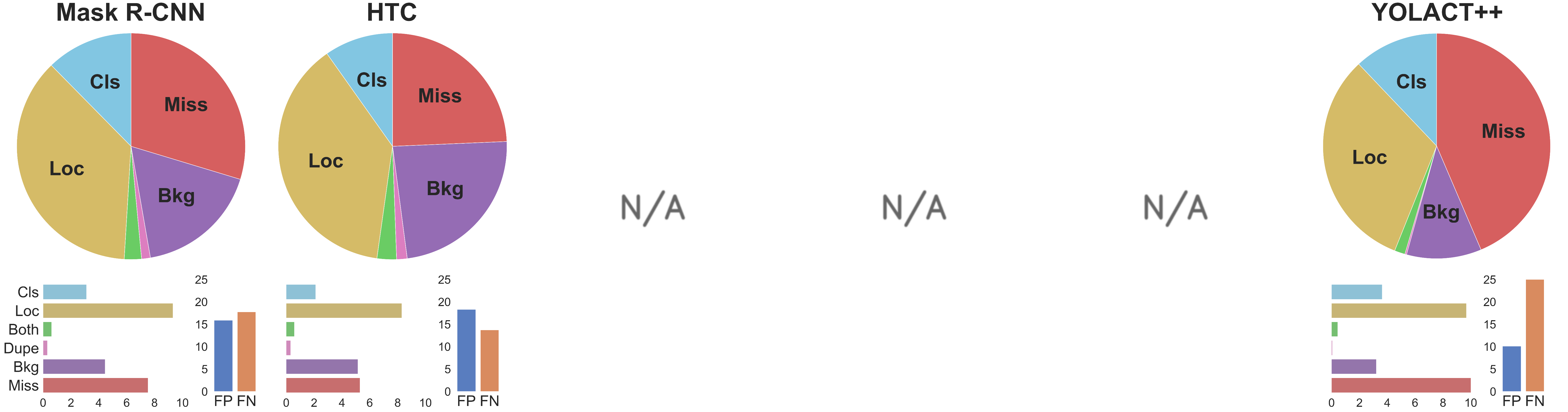} }}
    
    \caption{\textbf{Summary of errors on COCO.} The same as Fig.~3 in the main paper but with instance segmentation included. }
    
    \label{sup:fig:coco_summary}
\end{figure}

In Fig.~\ref{sup:fig:coco_summary} we show the summary plots for COCO instance segmentation that didn't make it into the main paper. For convenience, we also reproduce the detection results from the main paper.

\section{More Misannotated COCO Ground Truth}

As discussed in the main section, we find that a surprising number of the most confident errors are due to misannotated ground truth. In our study, we looked at the top 100 most confident errors in each error type (using a uniform random sample for Missed GT since those errors don't have an associated prediction) and pick out all the missannotated and ambiguously annotated examples from them. An important point to note is that COCO doesn't require their annotators to annotate all instances in an image. Where they don't annotate something, they're supposed to mark the whole area as a ``crowd'' annotation (i.e., for each crowd, they need only annotate a few instances and then anything else in the crowd annotation during evaluation will be ignored). We find that issues with crowd annotations are very common in COCO, such as not including a crowd annotation when there should be one or drawing a crowd annotation but then not marking the flag for crowd annotations.

This leads us to seperate all the misannotations into 3 categories: missing crowd label (i.e., the crowd annotation existed, but it wasn't flagged as a crowd annotation and thus got treated like a regular annotation), bad annotation (e.g., wrong class, box drawn incorrectly, or the GT didn't exist when it should have), and ambiguous (otherwise questionable annotations such as action figures not annotated as people, reflections of object not annotated, etc.).

We summarize our findings of the top 100 Mask R-CNN box errors for each error category in Tab.~\ref{sup:tab:misannotations}. Localization, both, and background errors all have a worrying number of misannotated GT, with the both error type having a whole two thirds of the 100 most confident as misannotated! Furthermore, simply forgetting to mark a crowd box as a crowd is a surprisingly common mistake that causes localization errors. This suggests that very simple steps can be taken to improve the quality of these annotations (just fix this mislabeled crowds and draw crowds around those that don't have them). This might be a good idea for future work to pursue.

\begin{table}[tb]
    \newcommand{\spacer}{\phantom{aa}}
    \newcommand{\halfspacer}{\phantom{a}}
    
    \centering
    \caption{{\bf Distribution of Misannotations on COCO.} We sample the top 100 errors from each error type (and randomly for missed) and bin the misannotations we found into one of three categories. Because these are the most confident examples, they have a very large effect on overall $mAP$. }
    \begin{smalltable}{r r r r r r} \toprule
    & Cls & Loc & Both & Bkgnd & Missed \\\midrule
    Crowd Flag     & 1 & 22 &  1 &  3 &  0 \\
    Bad Annotation & 1 &  8 & 36 & 34 &  8 \\
    Ambiguous      & 5 &  2 & 29 & 20 &  3 \\\midrule
    Total          & 7 & 32 & 66 & 57 & 11 \\
    \bottomrule
    \end{smalltable}\\
    \label{sup:tab:misannotations}
\end{table}

Another crucial note is that these missannotations exist in the training set too. This means for instance that all the boxes that should be marked as crowds but aren't are being used in our models as training examples. Qualitatively, a common error detectors make is when they lump two or more objects into the same detection. This type of error isn't exclusive to one method (we've observed it in FCIS, Mask R-CNN, and YOLACT, which all are vastly different architectures). Perhaps this type of error is caused by bad training data. Certainly this type of misannotation seems very common, so we can't really confidently pin those types of errors as our detectors' fault.

Thus as a meta point, it's really important that we be careful about how much we trust a dataset. Many errors could actually be a dataset's fault and not a model's fault, but it's not very common to really explore the dataset when designing new architectures and testing on those datasets. We urge researchers in machine learning to not treat their datasets as black box benchmarks, since in many aspects the dataset matters as much as the method.

\section{Using Ignored Detections for Error Computation}

 \begin{figure}[t]
    \centering

    \includegraphics[width=.6\linewidth]{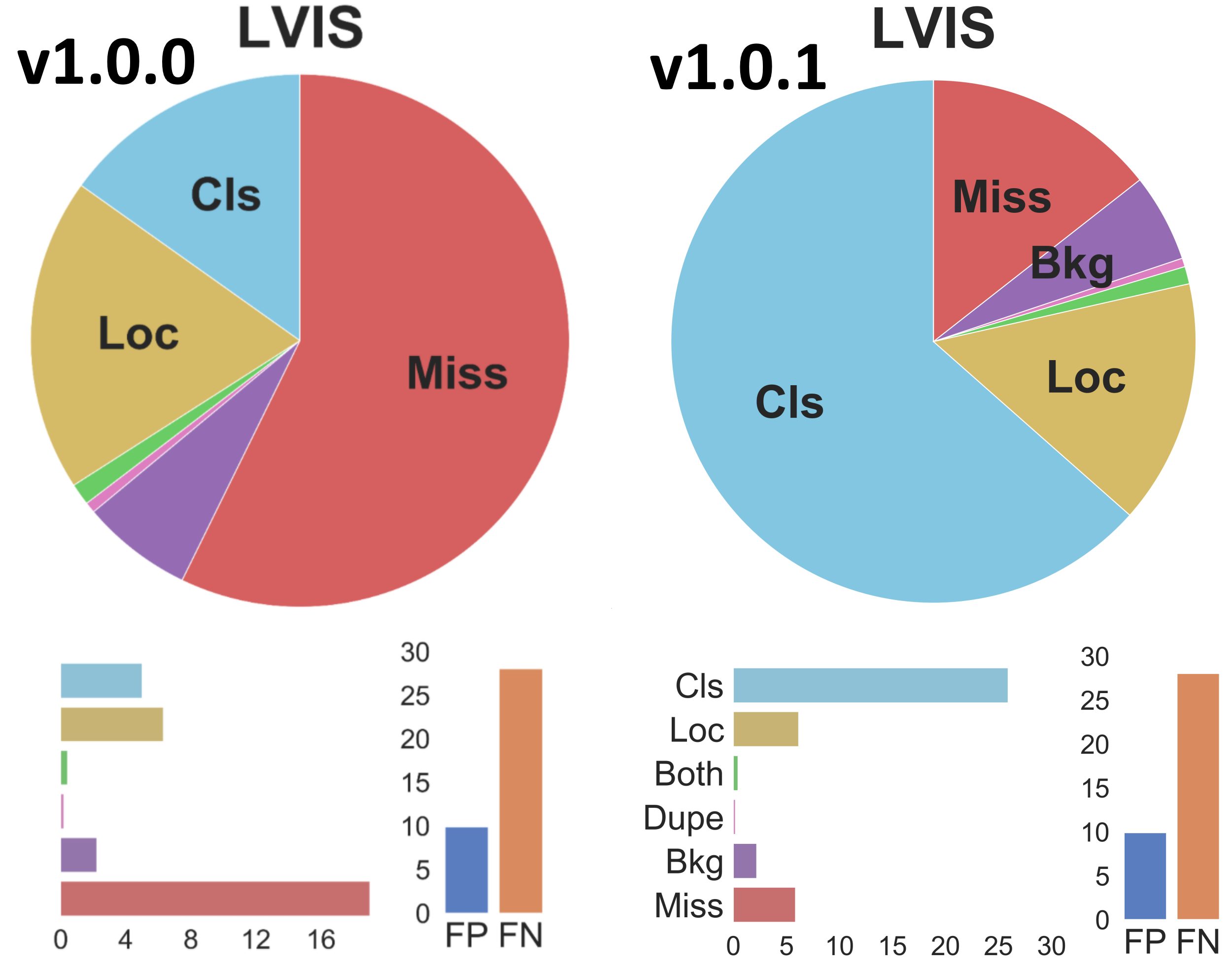}

    \caption{\textbf{Changes in Error Calculation.} In v1.0.1 we fixed an oversight that caused detections ignored in mAP calculation to also be ignored for error calculation. Since these ignored detections can still contribute to fixed mAP, we've solved this issue in v1.0.1 of the toolkit. This greatly affects results on LVIS, replacing almost all the missed error with classification error, but doesn't affect results on other datasets very much at all (since LVIS makes heavy use of ignoring detections). }
    \label{sup:fig:ignored_error}
\end{figure}

In the ECCV2020 version of this paper, we had an oversight in error computation that caused conclusions on LVIS \cite{gupta2019lvis} to be wrong. During mAP calculation on LVIS, most detections are simply ignored because they correspond to classes that aren't guaranteed to be in the image. Since these detections are ignored, they don't affect the mAP of the model on LVIS. However, it is still possible that fixing mistakes in those ignored detections would improve mAP. Namely, since most misclassifications are either not in the image or marked as not exhaustively annotated.

In v1.0.0 of the toolkit this would cause these misclassification errors to be marked as missed detections (since there was no non-ignored detection avaiable to match to the ground truth). In v1.0.1, we've fixed this issue which changed most of the missed ground truth error on LVIS to classification error. The resulting change is shown in Fig.~\ref{sup:fig:ignored_error}. The v1.0.1 result gives us a much more optimistic view of detectors on LVIS, since the primary goal of the dataset is to provide a classification challenge, not a recall challenge.

As for other datasets, including ignored detections in error calculation doesn't change the results that much (since those datasets don't make that heavy use of ignored regions). On COCO, for instance, the change in $\Delta AP$ is less than 0.4 for all error types.

%% file: main.bbl
\begin{thebibliography}{10}
\providecommand{\url}[1]{\texttt{#1}}
\providecommand{\urlprefix}{URL }
\providecommand{\doi}[1]{https://doi.org/#1}

\bibitem{COCOtoolkit}
{COCO} Analysis Toolkit: \url{http://cocodataset.org/\#detection-eval},
  accessed: 2020-03-01

\bibitem{bolya2019yolact++}
Bolya, D., Zhou, C., Xiao, F., Lee, Y.J.: Yolact++: Better real-time instance
  segmentation. arXiv:1912.06218  (2019)

\bibitem{bolya2019yolact}
Bolya, D., Zhou, C., Xiao, F., Lee, Y.J.: Yolact: real-time instance
  segmentation. In: ICCV (2019)

\bibitem{borji2019empirical}
Borji, A., Iranmanesh, S.M.: Empirical upper-bound in object detection and
  more. arXiv:1911.12451  (2019)

\bibitem{chen2019htc}
Chen, K., Pang, J., Wang, J., Xiong, Y., Li, X., Sun, S., Feng, W., Liu, Z.,
  Shi, J., Ouyang, W., et~al.: Hybrid task cascade for instance segmentation.
  In: CVPR (2019)

\bibitem{cordts2016cityscapes}
Cordts, M., Omran, M., Ramos, S., Rehfeld, T., Enzweiler, M., Benenson, R.,
  Franke, U., Roth, S., Schiele, B.: The cityscapes dataset for semantic urban
  scene understanding. In: CVPR (2016)

\bibitem{divvala2009context}
Divvala, S.K., Hoiem, D., Hays, J.H., Efros, A.A., Hebert, M.: An empirical
  study of context in object detection. In: CVPR (2009)

\bibitem{dollar2009caltechpeds}
Doll\'ar, P., Wojek, C., Schiele, B., Perona, P.: Pedestrian detection: A
  benchmark. In: CVPR (2009)

\bibitem{dong2017tumor}
Dong, H., Yang, G., Liu, F., Mo, Y., Guo, Y.: Automatic brain tumor detection
  and segmentation using u-net based fully convolutional networks. In: MIUA
  (2017)

\bibitem{everingham2010pascal}
Everingham, M., Van~Gool, L., Williams, C.K., Winn, J., Zisserman, A.: The
  pascal visual object classes (voc) challenge. IJCV  (2010)

\bibitem{girshick2014rcnn}
Girshick, R., Donahue, J., Darrell, T., Malik, J.: Rich feature hierarchies for
  accurate object detection and semantic segmentation. In: CVPR (2014)

\bibitem{gupta2019lvis}
Gupta, A., Dollar, P., Girshick, R.: Lvis: A dataset for large vocabulary
  instance segmentation. In: CVPR (2019)

\bibitem{he2017maskrcnn}
He, K., Gkioxari, G., Doll{\'a}r, P., Girshick, R.: Mask r-cnn. In: ICCV (2017)

\bibitem{hoiem2012diagnosing}
Hoiem, D., Chodpathumwan, Y., Dai, Q.: Diagnosing error in object detectors.
  In: ECCV (2012)

\bibitem{hosang2014proposals}
Hosang, J., Benenson, R., Schiele, B.: How good are detection proposals,
  really? In: BMVC (2014)

\bibitem{huang2019msrcnn}
Huang, Z., Huang, L., Gong, Y., Huang, C., Wang, X.: Mask scoring r-cnn. In:
  CVPR (2019)

\bibitem{kabra2015neighbors}
Kabra, M., Robie, A., Branson, K.: Understanding classifier errors by examining
  influential neighbors. In: CVPR (2015)

\bibitem{li2019tridentnet}
Li, Y., Chen, Y., Wang, N., Zhang, Z.: Scale-aware trident networks for object
  detection. In: ICCV (2019)

\bibitem{lin2017retinanet}
Lin, T.Y., Goyal, P., Girshick, R., He, K., Doll{\'a}r, P.: Focal loss for
  dense object detection. In: CVPR (2017)

\bibitem{lin2014coco}
Lin, T.Y., Maire, M., Belongie, S., Hays, J., Perona, P., Ramanan, D.,
  Doll{\'a}r, P., Zitnick, C.L.: Microsoft coco: Common objects in context. In:
  ECCV (2014)

\bibitem{liu2016ssd}
Liu, W., Anguelov, D., Erhan, D., Szegedy, C., Reed, S., Fu, C.Y., Berg, A.:
  Ssd: Single shot multibox detector. In: ECCV (2016)

\bibitem{pepik2015holding}
Pepik, B., Benenson, R., Ritschel, T., Schiele, B.: What is holding back
  convnets for detection? In: GCPR (2015)

\bibitem{tian2019fcos}
Tian, Z., Shen, C., Chen, H., He, T.: Fcos: Fully convolutional one-stage
  object detection. In: ICCV (2019)

\bibitem{zhu2015diagnosing}
Zhu, H., Lu, S., Cai, J., Lee, Q.: Diagnosing state-of-the-art object proposal
  methods. arXiv:1507.04512  (2015)

\end{thebibliography}
